\documentclass[letterpaper]{article} 
\usepackage{aaai23}  
\usepackage{times}  
\usepackage{helvet}  
\usepackage{courier}  
\usepackage[hyphens]{url}  
\usepackage{graphicx} 
\usepackage{natbib}  
\usepackage{caption} 
\usepackage{algorithm}
\usepackage{algorithmic}

\usepackage{xspace}
\usepackage{graphicx}
\usepackage{amsmath}
\usepackage{algorithm}
\usepackage{algorithmic}
\usepackage{wrapfig}
\usepackage{makecell}
\usepackage{subfigure}
\usepackage{relsize}
\usepackage{threeparttable}

\usepackage{booktabs}
\usepackage{xcolor}
\usepackage{amsfonts}

\newcommand{\NHL}[1]{\underline{\textbf{#1}}}
\newcommand{\ourmethod}{GREET\xspace}

\usepackage{newfloat}
\usepackage{listings}
\DeclareCaptionStyle{ruled}{labelfont=normalfont,labelsep=colon,strut=off} 
\floatstyle{ruled}
\newfloat{listing}{tb}{lst}{}
\floatname{listing}{Listing}
\title{Beyond Smoothing: Unsupervised Graph Representation Learning\\ with Edge Heterophily Discriminating}
\author {
    Yixin Liu\textsuperscript{\rm 1}, 
    Yizhen Zheng\textsuperscript{\rm 1}, 
    Daokun Zhang\textsuperscript{\rm 1}, 
    Vincent CS Lee\textsuperscript{\rm 1}, 
    Shirui Pan\textsuperscript{\rm 2}
}
\affiliations {
    \textsuperscript{\rm 1} Monash University, Australia; 
    \textsuperscript{\rm 2} Griffith University, Australia\\
    \{yixin.liu, yizhen.zheng1, daokun.zhang, vincent.cs.lee\}@monash.edu, s.pan@griffith.edu.au
}

\usepackage{bibentry}

\begin{document}

\maketitle

\begin{abstract}
Unsupervised graph representation learning (UGRL) has drawn increasing research attention and achieved promising results in several graph analytic tasks. Relying on the homophily assumption, existing UGRL methods tend to smooth the learned node representations along all edges, ignoring the existence of heterophilic edges that connect nodes with distinct attributes. 
As a result, current methods are hard to generalize to heterophilic graphs where dissimilar nodes are widely connected, and also vulnerable to adversarial attacks. 
{
To address this issue, we propose a novel unsupervised \NHL{G}raph \NHL{R}epresentation learning method with \NHL{E}dge h\NHL{E}terophily discrimina\NHL{T}ing (GREET) which learns representations by discriminating and leveraging homophilic edges and heterophilic edges. 
To distinguish two types of edges, we build an edge discriminator that infers edge homophily{/heterophily} from feature and structure information.} 
{We train the edge discriminator in an unsupervised way through minimizing the {crafted} pivot-anchored ranking loss, with randomly sampled node pairs acting as pivots. Node representations are learned through contrasting the dual-channel encodings obtained from the discriminated homophilic and heterophilic edges. With an effective interplaying scheme,} {edge discriminating and representation learning can mutually {boost} each other during the training phase.} We conducted extensive experiments on 14 benchmark datasets and multiple learning scenarios to demonstrate the superiority of GREET.

\end{abstract}

\section{Introduction} \label{sec:intro}
Unsupervised graph representation learning (UGRL) aims to learn low-dimensional representations from graph-structured data without costly labels~\cite{gae_kipf2016variational,sage_hamilton2017inductive,dgi_velivckovic2019deep}. Such representations can benefit myriads of downstream tasks, including node classification, link prediction and node clustering~\cite{zhang2022linkpred,arga_pan2018adversarially}. In recent years, UGRL has attracted increasing research attention and been applied to diverse applications, such as recommender systems~\cite{recon_ge2020graph}, traffic prediction~\cite{traffic_jenkins2019unsupervised}, and drug molecular modeling~\cite{drug_wang2021multi}.

\begin{figure*}[!t]
 \centering
 \vspace{-0.4cm}
 \subfigure[Cora (homophilic dataset)]{
   \includegraphics[height=0.17\textwidth]{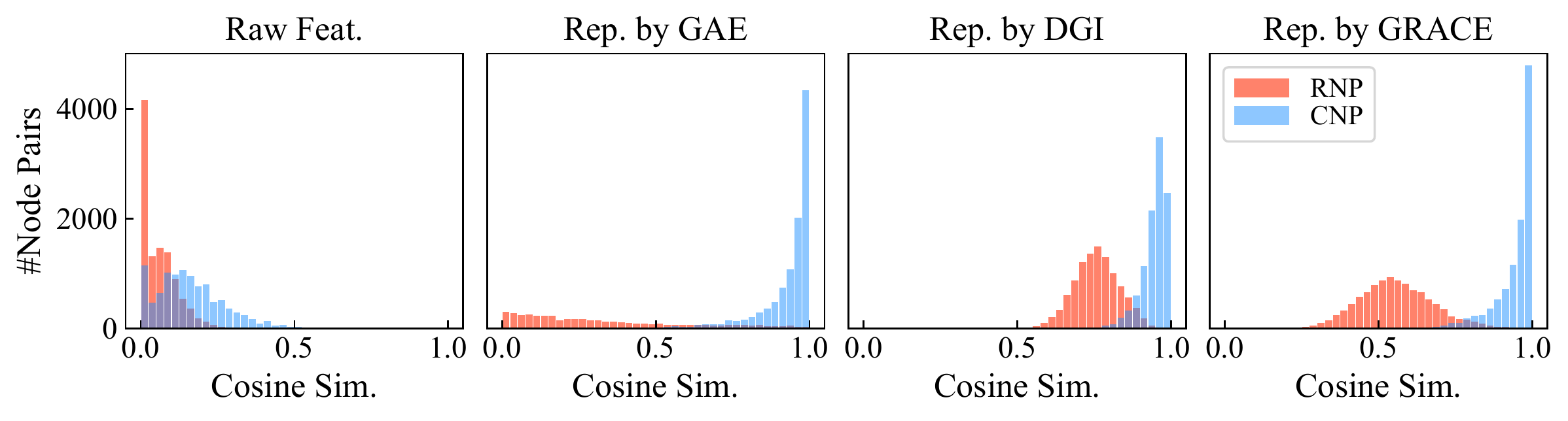}
   \label{subfig:sim_cora}
 } 
 \hspace{-0.2cm}
 \subfigure[Texas (heterophilic dataset)]{
   \includegraphics[height=0.17\textwidth]{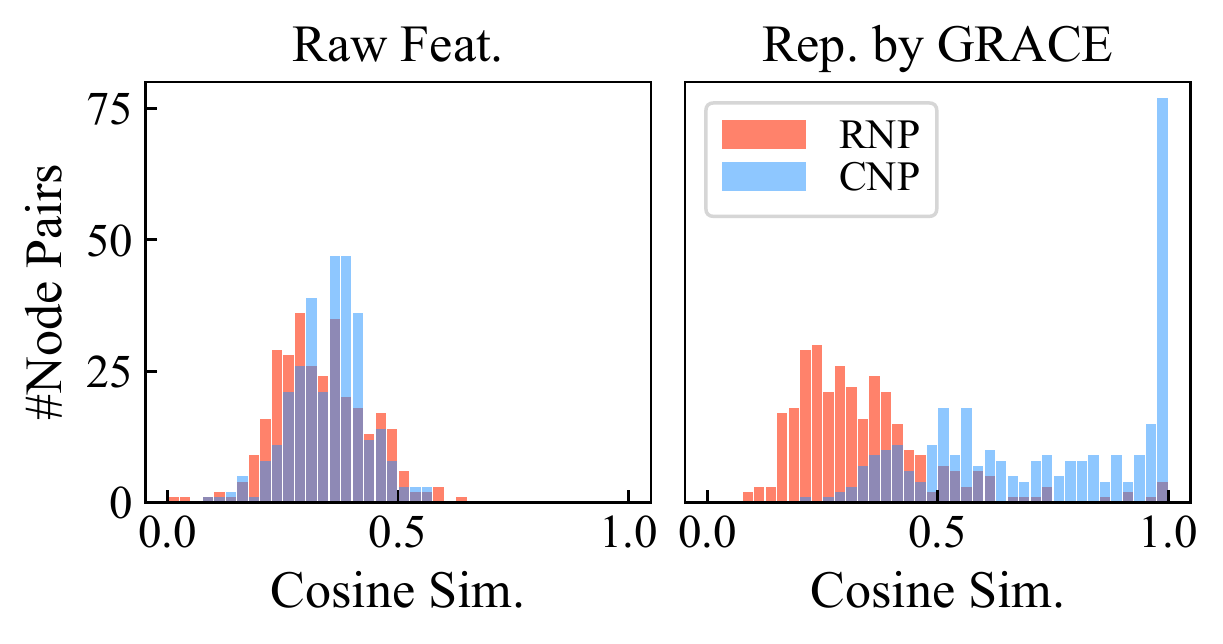}
   \label{subfig:sim_texas}
 }
 \vspace{-0.2cm}
 \caption{The distribution of pair-wise cosine similarity of connected node pairs~(CNP) and randomly sampled node pairs~(RNP). (a) The similarity w.r.t. raw features and representations learned by three UGRL methods (i.e., GAE \cite{gae_kipf2016variational}, DGI \cite{dgi_velivckovic2019deep}, and GRACE \cite{grace_zhu2020deep}) on Cora dataset. (b) The similarity w.r.t. raw features and representations learned by GRACE \cite{grace_zhu2020deep} on Texas dataset.}
 \vspace{-0.2cm}
 \label{fig:sim_intro}
\end{figure*}

Most of the UGRL methods are designed based on the homophily assumption, {i.e.}, linked nodes tend to share similar attributes with each other~\cite{assu1_hamilton2020graph}. Depending on such an assumption, they utilize low-pass filter-like graph neural networks (GNNs)~\cite{sgc_wu2019simplifying, gcn_kipf2017semi, lowpass_li2019label} to smooth the representations {of} adjacent nodes. Meanwhile, to provide supervision signals for representation learning, they often use objectives that preserve local smoothness, i.e., encouraging the representations of nodes within the same edge~\cite{gae_kipf2016variational, gmi_peng2020graph}, random walk~\cite{sage_hamilton2017inductive, perozzi2014deepwalk}, or subgraph~\cite{gcc_qiu2020gcc} to have higher similarity to each other. As a result, all the nodes are forced to have similar representations to their neighbors. 
As shown in Fig.~\ref{subfig:sim_cora}, all the connected nodes are pushed to be closer in the representation space, even if some of them have {moderate feature} similarities {that} {are comparable to} randomly sampled node pairs.  \looseness-1

With such a representation {smoothing mechanism}, existing UGRL methods cannot always generate optimal representations, since real-world graph data are likely to violate the homophily assumption~\cite{h2gcn_zhu2020beyond,fagcn_bo2021beyond}. Actually, {apart from} \textit{homophilic edges} that connect similar nodes, {there widely exist} \textit{heterophilic edges} in real-world graphs, {which otherwise} link dissimilar nodes. 
{Take social networks as an example, a person may have the same hobby with some friends, but also have quite different interests with other friends.} 
Moreover, in {a special type of graphs named} heterophilic graphs (e.g., Web Knowledge Base graph Texas, as shown in Fig.~\ref{subfig:sim_texas}), heterophilic edges are far more than homophilic ones~\cite{geom_pei2020geom}. Besides, graphs in reality inevitably contain noisy links caused by the uncertain {factors} in real-world systems or even adversarial attacks~\cite{hgsl_zhao2021heterogeneous, chang2021not}. In this situation, smoothing the information along all edges can lead to indistinguishable node representations, hindering the generalization ability and robustness of UGRL methods.  \looseness-1

Considering the existence of homophilic and heterophilic edges, a natural question can be raised: \textit{(Q1) Is it possible to distinguish between two types of edges {in an unsupervised manner}?} 
In supervised scenarios, {edge discriminating can be designed as a built-in component in node classification models, with labels providing direct evidence~\cite{zhao2021data}.} Nevertheless, in unsupervised scenarios, {it is hard to discriminate edge types through only node features and graph structure, especially for the widely existing edges that connect node pairs with moderate similarities.} {Even though a workable unsupervised edge discriminating strategy can be developed, we have to be faced with} a follow-up challenge:
\textit{(Q2) {How to effectively couple edge discriminating with representation learning into an integrated UGRL model?}} On the one hand, {for learning informative node representations, we have to make the best of the discriminated homophilic and heterophilic edges, which inevitably involve some errors. On the other hand, we shall find a good interplaying scheme to make edge discriminating and representation learning mutually boost each other.}

To answer both of the aforementioned questions, in this paper, we propose {a novel unsupervised \NHL{G}raph \NHL{R}epresentation learning method with \NHL{E}dge h\NHL{E}terophily discrimina\NHL{T}ing (GREET).}  Our theme is to {devise an effective unsupervised edge discriminating component, and integrate it with a dual-channel graph encoding component into an end-to-end UGRL framework}. More concretely, to answer \textit{(Q1)}, {for edge discriminating, we design a trainable edge discriminator with node features and structural encodings, and learn it in an unsupervised manner. To {provide} robust supervision, we design a pivot-anchored ranking loss, which uses the similarities of random node pairs as pivots, and evaluates the pivot-relative offside extent delivered by the end-node similarities of discriminated homophilic and heterophilic edges. Through minimizing the pivot-anchored ranking loss, we can effectively alleviate the misleading supervision caused by the ambiguous homophilic/heterophilic edges with moderate similarities.} To answer \textit{(Q2)}, {with the discriminated homophilic and heterophilic edges, we design a dual-channel graph encoding component to generate node representations, with each channel playing an essential role on the corresponding edge type view. A robust cross-channel contrasting mechanism is developed to learn informative node representations.} {To make the representation learning well integrated with edge discriminating, we use the learned node representations to measure node similarities for training the edge discriminator. Performed with a closed-loop interplay, edge discriminating and representation learning can increasingly promote each other during the model training phase.} {We conduct extensive experiments on 14 benchmark datasets, and the results demonstrate the superior effectiveness and robustness of our proposed \ourmethod over state-of-the-art methods.} \looseness-1

\section{Related Work} \label{sec:rw}
\vspace{-1mm}
In this section, we briefly review two related research directions. A more detailed review is available in Appendix A.

\noindent \textbf{Graph Neural Networks} (GNNs) aim to model graph-structured data via graph convolution based on spectral theory~\cite{estrach2014spectral,gcn_kipf2017semi, sgc_wu2019simplifying} or spatial information aggregation~\cite{sage_hamilton2017inductive,gat_velivckovic2018graph,gin_xu2019how}. From the perspective of graph signal processing, most GNNs can be viewed as low-pass graph filters that smooth features over graph topology, resulting in similar representations between adjacent nodes~\cite{revisit_nt2019revisiting,sgc_wu2019simplifying}. This property, unfortunately, hinders the capability of GNNs in heterophilic graphs where dissimilar nodes tend to be connected~\cite{linkx_lim2021large}. Some recent GNNs try to tackle this issue with novel designs~\cite{zheng2022graph}, such as advanced aggregation functions~\cite{geom_pei2020geom,fagcn_bo2021beyond,ugcn_jin2021universal} and network architecture~\cite{h2gcn_zhu2020beyond,gpr_chien2021adaptive,wrgnn2021}. However, they mainly focus on semi-supervised learning, leaving works on unsupervised learning unexplored. \looseness-2

\noindent \textbf{Unsupervised Graph Representation Learning} (UGRL) aims to learn low-dimensional node representations on graphs. Early methods learn by maximizing the representation similarity among proximal nodes (within the same random walk~\cite{perozzi2014deepwalk,node2vec_grover2016node2vec} or same edge~\cite{gae_kipf2016variational,arga_pan2018adversarially}). 
Recent efforts apply contrastive learning~\cite{simclr_chen2020simple} to UGRL, which optimize models by maximizing the agreement between two augmented graph views~\cite{mvgrl_hassani2020contrastive, gca_zhu2021graph,grace_zhu2020deep, dgi_velivckovic2019deep,gcc_qiu2020gcc,liu2022graph_ssl_survey,liu2022towards,liu2022good}. However, due to their low-pass filtering GNN encoders~\cite{gcn_kipf2017semi,sgc_wu2019simplifying} and smoothness-increasing learning objectives~\cite{perozzi2014deepwalk, gae_kipf2016variational, gmi_peng2020graph,gcc_qiu2020gcc}, existing UGRL methods tend to smooth the representations of every connected node pair, leading to their sub-optimal performance, especially on noisy or heterophilic graphs. Another line of methods aims to learn representations from multiple edge-based pre-defined graph views~\cite{qu2017attention, dmgi_park2020unsupervised,sign3_li2020learning, sign4_lee2020asine}. Differently, \ourmethod learns homophilic and heterophilic from graphs with a single edge type, which is more challenging.

\vspace{-1mm}
\section{Problem Definition} \label{sec:def}
\textbf{Notation.} 
Define a graph as $\mathcal{G}=(\mathcal{V}, \mathcal{E})$, where $\mathcal{V} = \{v_1, \cdots, v_{n} \}$ represents the node set and $\mathcal{E} \subseteq \mathcal{V} \times \mathcal{V}$ represents the edge set. The edge connecting node $v_i$ and $v_j$ is denoted as $e_{i,j}$. The numbers of nodes and edges are denoted as $|\mathcal{V}|=n$ and $|\mathcal{E}|=m$, respectively. Let $\mathbf{X} \in \mathbb{R}^{n \times d_f}$ denote the feature matrix, where the $i$-th row $\mathbf{x}_i$ represents the $d_f$-dimensional feature vector of node $v_i$. We denote the adjacency matrix of $\mathcal{G}$ as $\mathbf{A} \in \mathbb{R}^{n \times n}$, where $\mathbf{A}_{i j}=1$ if $e_{i,j} \in \mathcal{E}$ and $\mathbf{A}_{i j}=0$ otherwise. Using the feature matrix and adjacency matrix, the graph can also be written as $\mathcal{G}=(\mathbf{A}, \mathbf{X})$. 
The symmetric normalized adjacency matrix is denoted as $\tilde{\mathbf{A}} = \mathbf{D}^{-1 / 2} \mathbf{A} \mathbf{D}^{-1 / 2}$, where $\mathbf{D}$ is the diagonal degree matrix such that $\mathbf{D}_{i i}=\sum_{j} \mathbf{A}_{i j}$. The Laplacian matrix of the graph is defined as $\mathbf{L} = \mathbf{D} - \mathbf{A}$, and the symmetric normalized Laplacian matrix is $\tilde{\mathbf{L}} = \mathbf{I} - \tilde{\mathbf{A}}$.

\noindent \textbf{Problem Description.} 
We aim to solve the node-level unsupervised graph representation learning (UGRL) problem. The objective is to learn a representation {mapping} function $\mathcal{F}: \mathbb{R}^{n \times d_f} \times \mathbb{R}^{n \times n} \rightarrow \mathbb{R}^{n \times d_r}$ to {compute} a representation matrix $\mathcal{F}(\mathbf{X}, \mathbf{A}) = \mathbf{H}$, where the $i$-th row represents $\mathbf{h}_i$, the low-dimensional (i.e., $d_r \ll d_f$) representation of node $v_i$. These representations can be saved and utilized for downstream tasks, such as node classification.

\vspace{-1mm}
\section{Methodology} \label{sec:method}
\begin{figure*}[!t]
\centering
  \includegraphics[width=0.95\textwidth]{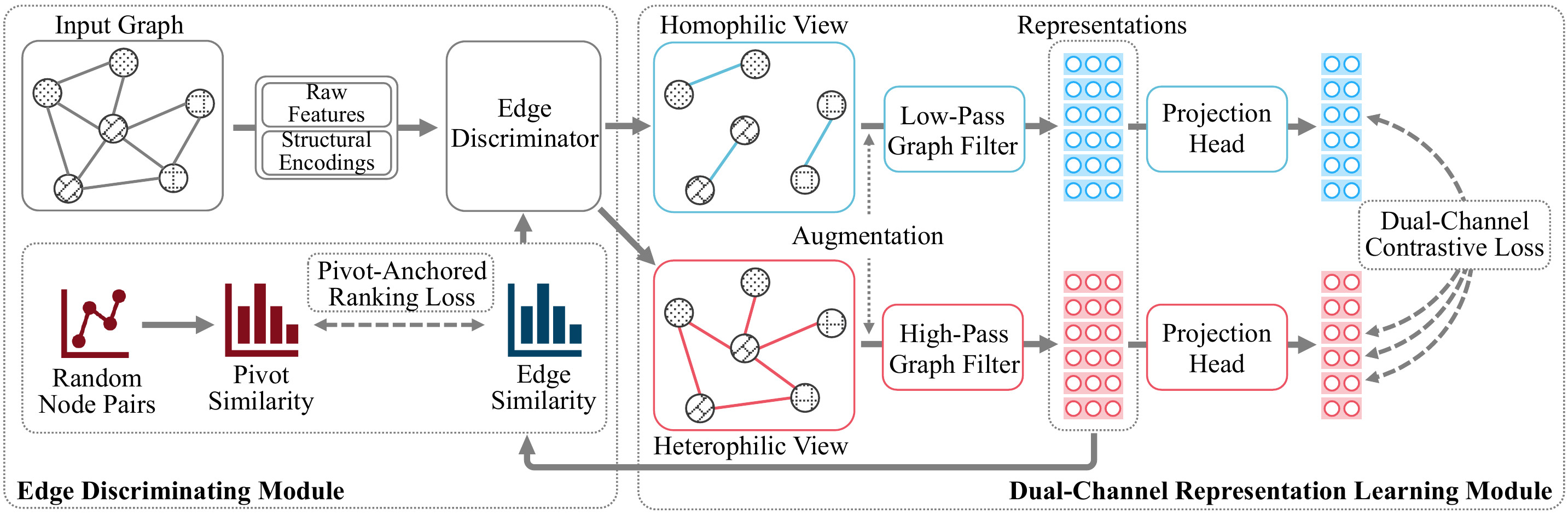}
  \vspace{-0.3cm}
  \caption{The overall pipeline of \ourmethod. In the \textit{edge discriminating module}, an edge discriminator is built to estimate the homophily probability of edges; in the \textit{dual-channel representation learning module}, {a} dual-channel {encoding component} generates node representations from {the discriminated} homophilic and heterophilic {edge} views. The edge discriminator and {encoding component} are trained by {minimizing} a pivot-anchored ranking loss and a robust dual-channel contrastive loss, respectively. \looseness-1
  }
  \label{fig:pipeline}
  \vspace{-0.3cm}
\end{figure*}

This section details our proposed method termed \ourmethod. As illustrated in Fig.~\ref{fig:pipeline}, \ourmethod is mainly composed of two components: \textit{edge discriminating module} that discriminates the homophilic and heterophilic edges, and \textit{dual-channel representation learning module} that leverages both types of edges to generate informative node representations. 
{In \textit{edge discriminating module}, the edges are {separated} into homophilic and heterophilic edges by an edge discriminator. In \textit{dual-channel representation learning module}, {node representations are constructed through applying} a dual-channel {encoding component} to perform graph convolution with low-pass and high-pass filters on the two dicriminated views respectively. The two main components are trained in a mutually {boosting} manner, with a pivot-anchored ranking loss used to train the edge discriminator, a robust dual-channel contrastive loss employed to learn informative node representations{,} and an effective alternating training {strategy} leveraged to make the the training on the two components increasingly reinforce each other.}

\subsection{Edge {Discriminating}} \label{subsec:discriminator}

In order to distinguish the homophilic and heterophilic edges, we construct an edge discriminator that learns to estimate the homophily probability of each edge. {Given two adjacent nodes, their homophily probability is essentially related to two critical factors: \textit{node features} which incorporate their semantic contents and \textit{structural characteristics} which reflect their topological roles}. 
A direct tool to integrate both structure and feature information is GNN \cite{gcn_kipf2017semi}; however, as we discussed before, GNN tends to smooth the features and then lose the heterophily-related knowledge. Thus, we consider an alternative solution, that is, modeling the structural characteristics with vectorial structural encoding (a.k.a. positional encoding)~\cite{rwpe_li2020distance, lspe_dwivedi2022graph, tan2023federated}. In this way, {the effective evidence can be preserved through concatenating node structural encodings with raw features.} {Specifically}, we employ a random walk diffusion process-based structural encoding~\cite{lspe_dwivedi2022graph} to {capture} structural information. The $d_s$-dimensional structural encoding $\mathbf{s}_i$ of node $v_i$ can be computed by $d_s$-step random walk-based graph diffusion:

\vspace{-0.1cm}
\begin{equation}
\label{eq:str_enc}
\mathbf{s}_{i}=\left[{\mathbf{T}}_{i i}, {\mathbf{T}}_{i i}^{2}, \cdots, {\mathbf{T}}_{i i}^{d_s}\right] \in \mathbb{R}^{d_s},
\end{equation}

\noindent where ${\mathbf{T}}=\mathbf{A}\mathbf{D}^{-1}$ is the random walk transition matrix. 
Taking the raw features and structural encodings as input, the edge discriminator {estimates} the homophily probability with two MLP layers:

\vspace{-0.1cm}
\begin{equation}
\label{eq:edge_discriminator}
\begin{aligned}
\mathbf{h}^{\prime}_i &= \mathrm{MLP}_{1}([\mathbf{x}_i \| \mathbf{s}_i]),\; \mathbf{h}^{\prime}_j = \mathrm{MLP}_{1}([\mathbf{x}_j \| \mathbf{s}_j]),\\
\theta_{i,j} &= \big(\mathrm{MLP}_{2}( [\mathbf{h}^{\prime}_i \|\mathbf{h}^{\prime}_j]) + \mathrm{MLP}_{2}( [\mathbf{h}^{\prime}_j \|\mathbf{h}^{\prime}_i])\big)\big/2,
\end{aligned}
\end{equation}

\noindent where $[\cdot\| \cdot]$ denotes the concatenation operation,  $\mathbf{h}^{\prime}_i$ is an intermediate embedding of node $v_i$, and $\theta_{i,j}$ is the {estimated} homophily probability for $e_{i,j}$. 
{To make the estimation not affected by the direction of edges, we apply the second MLP layer on the different orders of embedding concatenations.}

{{With the estimated} probability $\theta_{i,j}$, our purpose is to sample a binary homophily indicator ${w}_{{i,j}}$ {for edge $e_{i,j}$} from the Bernoulli distribution ${w}_{{i,j}} \sim \mathrm{Bernoulli}(\theta_{i,j})$,  with ${w}_{{i,j}}=1$ {indicating} homophily and ${w}_{{i,j}}=0$ {indicating} heterophily. However, the sampling is non-differentiable, making the discriminator hard to train. 
To address this issue, we approximate {the binary indicator} ${w}_{{i,j}}$ with Gumbel-Max reparametrization trick~\cite{gumbel1_jang2016categorical,gumbel2_maddison2016concrete}.} Specifically, the discrete homophily indicator ${w}_{{i,j}}$ is relaxed to a continuous homophily weight $\hat{w}_{{i,j}}$, which {is} computed by:

\vspace{-0.15cm}
\begin{equation}
\hat{w}_{{i,j}}= \mathrm{Sigmoid}\Big( \big( \theta_{i,j} +  \mathrm{log}\delta - \mathrm{log}(1-\delta) \big) / \tau_g \Big),
\end{equation}

\noindent where $\delta \sim \mathrm{Uniform}(0,1)$ is the {sampled} Gumbel random variate and $\tau_g > 0$ is the temperature hyper-parameter. {$\hat{w}_{{i,j}}$ tends to be sharper (closer to $0$ or $1$) with $\tau_g$ closer to $0$.}

{With the estimated edge homophily indicators, we discriminate the original graph $\mathcal{G}=(\mathbf{A}, \mathbf{X})$ into the homophilic and heterophilic graph views, $\mathcal{G}^{(hm)}=(\mathbf{A}^{(hm)}, \mathbf{X})$ and $\mathcal{G}^{(ht)}=(\mathbf{A}^{(ht)}, \mathbf{X})$. To make the edge discriminating trainable, instead of explicitly separating edges into homophilic and heterophilic categories, we assign a soft weight for each edge when it works on the homophilic/heterophilic view:}

\vspace{-0.15cm}
\begin{equation}
\mathbf{A}^{(hm)}_{i,j}=\hat{w}_{{i,j}},\;\mathbf{A}^{(ht)}_{i,j}=1-\hat{w}_{{i,j}},\;\mathrm{for}\;e_{i,j}\in\mathcal{E}.
\end{equation}

\subsection{Dual-Channel {Encoding}} \label{subsec:encoders}

{To learn informative representations from homophilic and heterophilic graph views, we design two different encoders that respectively perform low-pass and high-pass graph filtering on two distinct views. Our key motivation is to capture the shared information among similar nodes according to homophilic edges, while filtering out irrelevant information from dissimilar neighbors according to heterophilic edges.}

Concretely, {on} the homophilic view where similar nodes are connected to each other, we smooth the node features along the homophilic structure with a low-pass graph filter~\cite{dmp_yang2021diverse}. The low-pass filtering can capture the low-frequency information in graph signals and then preserve the shared knowledge of similar nodes. In \ourmethod, a simple low-pass GNN, SGC~\cite{sgc_wu2019simplifying}, serves as the low-pass graph filter, which is expressed as: \looseness-1

\begin{equation}
\mathbf{H}_0^{(hm)} = \mathrm{MLP}^{(hm)}(\mathbf{X}),\; \mathbf{H}_{l}^{(hm)} = \tilde{\mathbf{A}}^{(hm)} \mathbf{H}_{l-1}^{(hm)},
\end{equation}
\noindent where $\tilde{\mathbf{A}}^{(hm)}$ is the symmetric normalized homophilic adjacency matrix, $l \in \{1, \cdots, L\}$ is the layer index, and the final output $\mathbf{H}_L^{(hm)}$ {forms} the homophilic representation matrix $\mathbf{H}^{(hm)}$.

Different from the homophilic view, the heterophilic view contains edges connecting dissimilar nodes. In this view, smoothing features with low-pass filtering would confuse the information of different nodes and lead to the loss of discriminative node properties~\cite{fagcn_bo2021beyond}. To better leverage the heterophilic edges, we use high-pass graph filtering, the inverse operation of low-pass filtering, to sharpen the node features along edges and preserve high-frequency graph signals. In this way, the representations of dissimilar but connected nodes can be distinguished from each other. 
From the perspective of graph signal processing, graph Laplacian operator (i.e., multiplying the graph Laplacian matrix $\mathbf{L}$) has been proved to be effective to capture high-frequency components~\cite{hp_dong2021adagnn,hp_ma2021unified}. 
Accordingly, we design a simple GNN, namely Lap-SGC, to perform high-pass filtering on the heterophilic view:

\begin{equation}
\mathbf{H}_0^{(ht)} = \mathrm{MLP}^{(ht)}(\mathbf{X}),\; \mathbf{H}_{l}^{(ht)} = \tilde{\mathbf{L}}^{(ht)} \mathbf{H}_{l-1}^{(ht)},
\end{equation}

\noindent where {$l \in \{1, \cdots, L\}$}, and $\tilde{\mathbf{L}}^{(ht)} = \mathbf{I} - \alpha \tilde{\mathbf{A}}^{(ht)}$ is the symmetric normalized Laplacian matrix of heterophilic view {with} $\alpha$ {being} a hyper-parameter to control the strength of high-pass filtering. We take the final output of Lap-SGC {$\mathbf{H}^{(ht)}_{L}$} as the heterophilic representation matrix $\mathbf{H}^{(ht)}$.

Using {two distinct} encoders (i.e., SGC and Lap-SGC), now we {acquire} two groups of representations that {capture} low-frequency and high-frequency information respectively. {Final node representations are obtained through concatenating} the representations {constructed} from both views: $\mathbf{H} = [\mathbf{H}^{(hm)}\| \mathbf{H}^{(ht)}]${$\in\mathbb{R}^{n\times d_r}$ with its $i$-th row $\mathbf{h}_i\in\mathbb{R}^{d_r}$ being the representation vector of node $v_i$}. \looseness-1

\subsection{{Model Training}} \label{subsec:training}

{To jointly learn edge distinction and node representations, we carefully design the learning objectives and training {strategy} for \ourmethod. Specifically, we design a pivot-anchored ranking loss to train the edge discriminator, and construct a robust dual-channel contrastive loss to train the {representation} {encoders}. We also introduce an alternating training {strategy} to iteratively optimize two components.}

\noindent \textbf{Pivot-Anchored Ranking Loss.} 
The target of the edge discriminator is to distinguish between homophilic edges (connecting similar nodes) and heterophilic edges (connecting dissimilar nodes), which is not a trivial task in an unsupervised scenario. The main challenge is to find the boundary between ``similar'' and ``dissimilar''. 
{To this end, {we {propose to use} the randomly sampled node pairs as ``pivots'' for similarity measurement. Based on {them},} we introduce a {pivot-anchored} ranking loss to make sure that node pairs connected by homophilic edges are significantly more similar than the {pivot} node pairs, while the node pairs connected by heterophilic edges are markedly more different than the {pivot} node pairs. {{In detail}}, for each homophilic/heterophilic edge $e_{i,j}\in\mathcal{E}$, its homophilic {pivot-anchored} ranking loss $\mathcal{R}^{(hm)}(e_{i,j})$ and heterophilic {pivot-anchored} ranking loss $\mathcal{R}^{(ht)}(e_{i,j})$ are respectively defined as:}
\begin{equation}
\begin{aligned}
\mathcal{R}^{(hm)}(e_{i,j}) &= [s_{v_{i^{\prime}},v_{j^{\prime}}}-s_{e_{i,j}}+\gamma^{(hm)}]_{+},\;\\
\mathcal{R}^{(ht)}(e_{i,j}) &= [s_{e_{i,j}}-s_{v_{i^{\prime}},v_{j^{\prime}}}+\gamma^{(ht)}]_{+},
\end{aligned}
\end{equation}
{where $[x]_{+}=\max(x,0)$ denotes only the positive case of $x$ is considered; $s_{e_{i,j}}=\cos(\mathbf{h}_i, \mathbf{h}_j)$ and $s_{v_{i^{\prime}},v_{j^{\prime}}}=\cos(\mathbf{h}_{i^{\prime}}, \mathbf{h}_{j^{\prime}})$ are respectively the representation-level cosine similarity between the node pair connected by edge $e_{i,j}$ and the cosine similarity between two randomly sampled nodes $v_{i^{\prime}}$ and $v_{j^{\prime}}$; $\gamma^{(hm)}$ and $\gamma^{(ht)}$ are respectively the margin parameters for the homophilic and heterophilic {pivot-anchored} ranking losses, which {can} force the similarity difference to reach a significant level. }

\begin{table*}[t]
\vspace{-0.1cm}
\centering
\caption{Results in terms of classification accuracies (in percent $\pm$ standard deviation) on homophilic benchmarks. ``*'' indicates that results are borrowed from the original papers. OOM indicates Out-Of-Memory on a 24GB GPU. The best and runner-up results are highlighted with \textbf{bold} and \underline{underline}, respectively.} 
\label{tab:main_result_homo}
\vspace{-0.1cm}
\resizebox{0.95\textwidth}{!}{
\begin{tabular}{p{2.cm}|p{2.cm}<{\centering} p{2.cm}<{\centering} p{2.cm}<{\centering} p{2.cm}<{\centering} p{2.cm}<{\centering} p{2.cm}<{\centering} p{2.cm}<{\centering} p{2.cm}<{\centering}}
\toprule
Methods & Cora & CiteSeer & PubMed & Wiki-CS & Amz. Comp. & Amz. Photo & Co. CS & Co. Physics \\
\midrule
GCN*      & ${81.5}$ & ${70.3}$ & ${79.0}$ & $76.89{\scriptstyle\pm0.37}$ & $86.34{\scriptstyle\pm0.48}$ & $92.35{\scriptstyle\pm0.25}$ & $93.10{\scriptstyle\pm0.17}$ & $95.54{\scriptstyle\pm0.19}$  \\
GAT*      & ${83.0}$ & ${72.5}$ & ${79.0}$ & $77.42{\scriptstyle\pm0.19}$ & $87.06{\scriptstyle\pm0.35}$ & $92.64{\scriptstyle\pm0.42}$ & $92.41{\scriptstyle\pm0.27}$ & $95.45{\scriptstyle\pm0.17}$  \\
MLP      & $56.11{\scriptstyle\pm0.34}$ & $56.91{\scriptstyle\pm0.42}$ & $71.35{\scriptstyle\pm0.05}$ & $72.02{\scriptstyle\pm0.21}$ & $73.88{\scriptstyle\pm0.10}$ & $78.54{\scriptstyle\pm0.05}$ & $90.42{\scriptstyle\pm0.08}$ & $93.54{\scriptstyle\pm0.05}$  \\
\midrule
DeepWalk & $69.47{\scriptstyle\pm0.55}$ & $58.82{\scriptstyle\pm0.61}$ & $69.87{\scriptstyle\pm1.25}$ & $74.35{\scriptstyle\pm0.06}$ & $85.68{\scriptstyle\pm0.06}$ & $89.44{\scriptstyle\pm0.11}$ & $84.61{\scriptstyle\pm0.22}$ & $91.77{\scriptstyle\pm0.15}$  \\
node2vec & $71.24{\scriptstyle\pm0.89}$ & $47.64{\scriptstyle\pm0.77}$ & $66.47{\scriptstyle\pm1.00}$ & $71.79{\scriptstyle\pm0.05}$ & $84.39{\scriptstyle\pm0.08}$ & $89.67{\scriptstyle\pm0.12}$ & $85.08{\scriptstyle\pm0.03}$ & $91.19{\scriptstyle\pm0.04}$  \\
GAE      & $71.07{\scriptstyle\pm0.39}$ & $65.22{\scriptstyle\pm0.43}$ & $71.73{\scriptstyle\pm0.92}$ & $70.15{\scriptstyle\pm0.01}$ & $85.27{\scriptstyle\pm0.19}$ & $91.62{\scriptstyle\pm0.13}$ & $90.01{\scriptstyle\pm0.71}$ & $94.92{\scriptstyle\pm0.07}$  \\
VGAE     & $79.81{\scriptstyle\pm0.87}$ & $66.75{\scriptstyle\pm0.37}$ & $77.16{\scriptstyle\pm0.31}$ & $75.63{\scriptstyle\pm0.19}$ & $86.37{\scriptstyle\pm0.21}$ & $92.20{\scriptstyle\pm0.11}$ & $92.11{\scriptstyle\pm0.09}$ & $94.52{\scriptstyle\pm0.00}$  \\
\midrule
DGI      & $82.29{\scriptstyle\pm0.56}$ & $71.49{\scriptstyle\pm0.14}$ & $77.43{\scriptstyle\pm0.84}$ & $75.73{\scriptstyle\pm0.13}$ & $84.09{\scriptstyle\pm0.39}$ & $91.49{\scriptstyle\pm0.25}$ & $91.95{\scriptstyle\pm0.40}$ & $94.57{\scriptstyle\pm0.38}$  \\
GMI      & $82.51{\scriptstyle\pm1.47}$ & $71.56{\scriptstyle\pm0.56}$ & $79.83{\scriptstyle\pm0.90}$ & $75.06{\scriptstyle\pm0.13}$ & $81.76{\scriptstyle\pm0.52}$ & $90.72{\scriptstyle\pm0.33}$ & OOM & OOM  \\
MVGRL    & $\underline{83.03{\scriptstyle\pm0.27}}$ & $\underline{72.75{\scriptstyle\pm0.46}}$ & $79.63{\scriptstyle\pm0.38}$ & $77.97{\scriptstyle\pm0.18}$ & $87.09{\scriptstyle\pm0.27}$ & $92.01{\scriptstyle\pm0.13}$ & $91.97{\scriptstyle\pm0.19}$ & $95.53{\scriptstyle\pm0.10}$  \\
GRACE    & $80.08{\scriptstyle\pm0.53}$ & $71.41{\scriptstyle\pm0.38}$ & $80.15{\scriptstyle\pm0.34}$ & $79.16{\scriptstyle\pm0.36}$ & $87.21{\scriptstyle\pm0.44}$ & $92.65{\scriptstyle\pm0.32}$ & $92.78{\scriptstyle\pm0.23}$ & $95.39{\scriptstyle\pm0.32}$  \\
GCA      & $80.39{\scriptstyle\pm0.42}$ & $71.21{\scriptstyle\pm0.24}$ & $\mathbf{80.37{\scriptstyle\pm0.75}}$ & $\underline{79.35{\scriptstyle\pm0.12}}$ & $87.84{\scriptstyle\pm0.27}$ & $92.78{\scriptstyle\pm0.17}$ & $\underline{93.32{\scriptstyle\pm0.12}}$ & $\underline{95.87{\scriptstyle\pm0.15}}$  \\
BGRL     & $81.08{\scriptstyle\pm0.17}$ & $71.59{\scriptstyle\pm0.42}$ & $79.97{\scriptstyle\pm0.36}$ & $78.74{\scriptstyle\pm0.22}$ & $\mathbf{88.92{\scriptstyle\pm0.33}}$ & $\mathbf{93.24{\scriptstyle\pm0.29}}$ & ${93.26{\scriptstyle\pm0.36}}$ & $95.76{\scriptstyle\pm0.38}$  \\
\midrule 
\ourmethod & $\mathbf{83.81{\scriptstyle\pm0.87}}$ & $\mathbf{73.08{\scriptstyle\pm0.84}}$ & $\underline{80.29{\scriptstyle\pm1.00}}$ & $\mathbf{80.68{\scriptstyle\pm0.31}}$ & $\underline{87.94{\scriptstyle\pm0.35}}$ & $\underline{92.85{\scriptstyle\pm0.31}}$ & $\mathbf{94.65{\scriptstyle\pm0.18}}$ & $\mathbf{96.13{\scriptstyle\pm0.12}}$  \\
\bottomrule
\end{tabular}}
\vspace{-0.2cm}
\end{table*}

{Through performing expectation over all homophilic and heterophilic edges respectively, we can obtain the expected homophilic and heterophilic {pivot-anchored} ranking losses $\mathcal{L}_{r}^{(hm)}$ and $\mathcal{L}_{r}^{(ht)}$ as:}
\begin{equation}
\label{eq:wloss_2}
\begin{scriptsize}
\begin{aligned}
\mathcal{L}_{r}^{(hm)} = \mathop{\mathlarger{\mathbb{E}}}\limits_{e_{i,j}\sim\Pi(\hat{w}_{i,j})}\mathcal{R}^{(hm)}(e_{i,j})&=\frac{1}{\hat{W}^{(hm)}}\sum_{e_{i,j}\in\mathcal{E}}\hat{w}_{i,j}\mathcal{R}^{(hm)}(e_{i,j}),
\\
\mathcal{L}_{r}^{(ht)} = \mathop{\mathlarger{\mathbb{E}}}\limits_{e_{i,j}\sim\Pi(1-\hat{w}_{i,j})}\mathcal{R}^{(ht)}(e_{i,j})&=\frac{1}{\hat{W}^{(ht)}}\sum_{e_{i,j}\in\mathcal{E}}(1-\hat{w}_{i,j})\mathcal{R}^{(ht)}(e_{i,j}),
\end{aligned}
\end{scriptsize}
\end{equation}

{where $\Pi(\hat{w}_{i,j})$ and $\Pi(1-\hat{w}_{i,j})$ are the multinomial distributions parameterized by homophilic and heterophilic edge weights $\hat{w}_{i,j}$ and $1-\hat{w}_{i,j}$; $\hat{W}^{(hm)}$ and $\hat{W}^{(ht)}$ are the normalization terms which are respectively obtained by summing over all homophilic edge weights $\hat{w}_{i,j}$ and heterophilic edge weights $1-\hat{w}_{i,j}$. Through summing up the two expected losses, the overall {pivot-anchored} ranking loss to be minimized can be obtained as:}
\begin{equation}
\label{eq:wloss_3}
\mathcal{L}_{r} = \mathcal{L}_{r}^{(hm)} + \mathcal{L}_{r}^{(ht)}.
\end{equation}

\noindent \textbf{{Robust Dual-Channel} Contrastive Loss.} 
{To provide supervision for node representation learning, we use a robust contrasting mechanism {that forces our model to generate semantically consistent representations from two distinct graph views.} 
{More importantly, such a learning target} can effectively combat negative impacts caused by the small amounts of false discriminated edges, through implicitly using the generated representations to reconstruct feature-level similarity. Specially, for each node $v_i\in\mathcal{V}$, we first find another node $v_j$ within $v_i$'s top-$k$ similar node set measured by features, $\mathcal{N}_i = \mathrm{kNN}(v_i, k)$, then compare their representations across homophilic and heterophilic views. 
To provide a fair similarity measure, we project node representations delivered by two views into a latent space with two learnable MLPs respectively, then use the similarity between the projections in the latent space as node representation similarity measuring metric. Inspired by the InfoNCE contrastive loss~\cite{simclr_chen2020simple, grace_zhu2020deep}, }the robust cross-channel contrastive loss is defined as: \looseness-1

\begin{equation}
\label{eq:closs}
\begin{scriptsize}
\begin{aligned}
\mathcal{L}_{c} = - \frac{1}{n} \sum_{v_i \in \mathcal{V}}\bigg[\frac{1}{2|\mathcal{N}_i|} \sum_{v_j \in \mathcal{N}_i} \bigg(\log \frac{e^{\cos\left(\mathbf{z}^{(hm)}_i, \mathbf{z}^{(ht)}_j\right) / \tau_c}}{\sum_{v_k \in \mathcal{V} \backslash v_i} e^{\cos\left(\mathbf{z}^{(hm)}_i, \mathbf{z}^{(ht)}_k\right) / \tau_c}}\\
+\log \frac{e^{\cos\left(\mathbf{z}^{(ht)}_i, \mathbf{z}^{(hm)}_j\right) / \tau_c}}{\sum_{v_k \in \mathcal{V} \backslash v_i} e^{\cos\left(\mathbf{z}^{(ht)}_i, \mathbf{z}^{(hm)}_k\right) / \tau_c}}\bigg)\bigg],
\end{aligned}
\end{scriptsize}
\end{equation}

\noindent where $\mathbf{z}^{(hm)}_i$ and $\mathbf{z}^{(ht)}_i$ are respectively the projections of $v_i$'s homophilic-view and heterophilic-view representations in the latent space, $\tau_c$ is a temperature hyper-parameter for contrastive learning and $\cos(\cdot,\cdot)$ is cosine similarity. In practice, we calculate contrastive loss in an mini-batch manner \cite{simclr_chen2020simple} for large graph datasets. 

\noindent \textbf{{Training Strategy}.} 
In our method, the training of edge discriminating module {relies on node representations to measure node similarity}{;}  the representation learning module, in turn, generates representations from {the two views} delivered by edge discriminating. To train two components effectively, we utilize an alternating training strategy to {optimize} edge {discriminating} and node representation {learning alternately, with the two components boosting each other increasingly.} To sum up, the overall optimization objective can be written as $\mathcal{L} = \mathcal{L}_r + \mathcal{L}_c$. 
To {improve the generalization ability of} the representation learning module, we adopt data augmentation that increases the diversity of data for model training~\cite{graphcl_you2020graph}. {Concretely, }we employ two simple but effective augmentation strategies, feature masking and edge dropping~\cite{grace_zhu2020deep,graphcl_you2020graph}, to perturb the features and structures of both views. {We present the algorithmic description in Appendix~B.} \looseness-2

\noindent {\textbf{Scalability Extension.} To improve the scalability of \ourmethod, we introduce the following mechanisms. (1) For structural encoding computation, we use sparse matrix multiplication to calculate graph diffusion, which avoids $\mathcal{O}(n^3)$ time complexity. (2) To find similar nodes in $\mathcal{L}_c$, we use locality-sensitive approximation algorithm \cite{fatemi2021slaps} to obtain kNN nodes. (3) We calculate $\mathcal{L}_c$ within a mini-batch of nodes instead of all nodes, reducing the time complexity from $\mathcal{O}(n^2)$ to $\mathcal{O}(nb)$, where $b$ is the size of mini-batch. Note that the structural encodings and kNN nodes can be pre-computed at once before model training. Experiments on CoAuthor Physics dataset verify that \ourmethod is able to learn on graphs with over 34k nodes and 495k edges. Detailed complexity analysis of \ourmethod can be found in Appendix~C.
}

\vspace{-1mm}
\section{Experiments} \label{sec:exp}
\subsection{Experimental Settings} \label{subsec:exp_setting}

\textbf{Datasets.} 
We take transductive node classification as the downstream task to evaluate the effectiveness of the learned representations. Our experiments are conducted on 14 commonly used benchmark datasets, including 8 homophilic graph datasets (i.e., Cora, CiteSeer, PubMed, Wiki-CS, Amazon Computer, Amazon Photo, CoAuthor CS, and CoAuthor Physics ~\cite{sen2008collective,mernyei2020wiki,shchur2018pitfalls}) and 6 heterophilic graph datasets (i.e., Chameleon, Squirrel, Actor, Cornell, Texas, and Wisconsin~\cite{geom_pei2020geom}). {We split all datasets following the public splits~\cite{yang2016revisiting,gcn_kipf2017semi,geom_pei2020geom} or commonly used splits~\cite{gca_zhu2021graph,bgrl_thakoor2021large}.} The details of datasets are summarized in Appendix~D.

\noindent \textbf{Baselines.} 
We compare \ourmethod with three groups of baseline methods: (1) supervised/semi-supervised learning methods (i.e., GCN~\cite{gcn_kipf2017semi}, GAT~\cite{gat_velivckovic2018graph}, and MLP), (2) conventional UGRL methods (i.e., node2vec~\cite{node2vec_grover2016node2vec}, DeepWalk~\cite{perozzi2014deepwalk}, GAE, and VGAE~\cite{gae_kipf2016variational}), and (3) contrastive UGRL methods (i.e., DGI~\cite{dgi_velivckovic2019deep}, GMI~\cite{gmi_peng2020graph}, MVGRL~\cite{mvgrl_hassani2020contrastive}, GRACE~\cite{grace_zhu2020deep}, GCA~\cite{gca_zhu2021graph}, and BGRL~\cite{bgrl_thakoor2021large}). For the experiments on heterophilic graphs, we further consider 4 {heterophily-aware} GNNs as baselines, including Geom-GCN~\cite{geom_pei2020geom}, H2GCN~\cite{h2gcn_zhu2020beyond}, FAGCN~\cite{fagcn_bo2021beyond}, and GPR-GNN~\cite{gpr_chien2021adaptive}. {We further equip GRACE with FAGCN encoder (termed GRACE-FA) for comparison.}

\begin{table}[t]
\centering
\caption{Results in terms of classification accuracies (in percent) on heterophilic benchmarks. ``*'' indicates that results are borrowed from the original papers. The best and runner-up results are highlighted with \textbf{bold} and \underline{underline}, respectively. {Results with standard deviation are in Appendix G.}} 
\label{tab:main_result_hete}
\vspace{-0.1cm}
\resizebox{1.\columnwidth}{!}{
\begin{tabular}{l | cccccc}
\toprule
Methods & Cham. & Squ. & Actor & Cor. & Texas & Wis. \\
\midrule
GCN      & $59.63$ & $36.28$ & $30.83$ & $57.03$ & $60.00$ & $56.47$ \\
GAT      & $56.38$ & $32.09$ & $28.06$ & $59.46$ & $61.62$ & $54.71$ \\
MLP      & $46.91$ & $29.28$ & $35.66$ & $81.08$ & $81.62$ & $84.31$ \\
\midrule
Geom-GCN* & ${60.90}$ & ${38.14}$ & ${31.63}$ & ${60.81}$ & ${67.57}$ & ${64.12}$ \\
H2GCN*    & $59.39$ & $37.90$ & $\underline{35.86}$ & $\underline{82.16}$ & $\underline{84.86}$ & $\mathbf{86.67}$ \\
FAGCN    & $\underline{63.44}$ & $\underline{41.17}$ & $35.74$ & $81.35$ & $84.32$ & $83.33$ \\
GPR-GNN  & $61.58$ & $39.65$ & $35.27$ & $81.89$ & $83.24$ & $84.12$ \\
\midrule
DeepWalk & $47.74$ & $32.93$ & $22.78$ & $39.18$ & $46.49$ & $33.53$ \\
node2vec & $41.93$ & $22.84$ & $28.28$ & $42.94$ & $41.92$ & $37.45$ \\
GAE      & $33.84$ & $28.03$ & $28.03$ & $58.85$ & $58.64$ & $52.55$ \\
VGAE     & $35.22$ & $29.48$ & $26.99$ & $59.19$ & $59.20$ & $56.67$ \\
\midrule
DGI      & $39.95$ & $31.80$ & $29.82$ & $63.35$ & $60.59$ & $55.41$ \\
GMI      & $46.97$ & $30.11$ & $27.82$ & $54.76$ & $50.49$ & $45.98$ \\
MVGRL    & $51.07$ & $35.47$ & $30.02$ & $64.30$ & $62.38$ & $62.37$ \\
GRACE    & $48.05$ & $31.33$ & $29.01$ & $54.86$ & $57.57$ & $50.00$ \\
{GRACE-FA} & $52.68$ & $35.97$ & $32.55$ & $67.57$ & $64.05$ & $63.73$ \\
GCA      & $49.80$ & $35.50$ & $29.65$ & $55.41$ & $59.46$ & $50.78$ \\
BGRL     & $47.46$ & $32.64$ & $29.86$ & $57.30$ & $59.19$ & $52.35$ \\
\midrule 
\ourmethod & $\mathbf{63.64}$ & $\mathbf{42.29}$ & $\mathbf{36.55}$ & $\mathbf{85.14}$ & $\mathbf{87.03}$ & $\underline{84.90}$ \\
\bottomrule
\end{tabular}}
\vspace{-0.2cm}
\end{table}

\noindent \textbf{Experimental details. } 
For \ourmethod and the unsupervised baselines, we follow the evaluation protocol that is widely used in previous UGRL methods~\cite{grace_zhu2020deep, gca_zhu2021graph}. Specifically, in training phase, the model are trained without supervision; then, in evaluation phase, the learned representations are frozen and used to train and test with a logistic regression classifier. For the superivsed/semi-supervised baselines, we directly train the models in an end-to-end fashion where labels are available for representation learning, {and perform evaluation with the trained models}. For all datasets, we report the averaged test accuracy and standard deviation over 10 runs of experiments. We conduct grid search to choice the best hyper-parameters on validation set. We also search the best hyper-parameter for baseline methods during reproduction. Specific hyper-parameter settings and more implementation details are in Appendix~E. {The experiments that investigate the parameter sensitivity of \ourmethod are demonstrated in Appendix~F.} The code of \ourmethod is available at \url{https://github.com/yixinliu233/GREET}.

\subsection{Experimental Results}

\noindent \textbf{Performance Comparison. } 
The node classification results on 8 homophilic datasets and 6 heterophilic datasets are presented in Table~\ref{tab:main_result_homo} and Table~\ref{tab:main_result_hete}, respectively. First of all, we observe that \ourmethod outperforms all baseline methods in 10 out of 14 benchmarks and achieves the runner-up performance on the rest 4 benchmarks. The superior performance indicates the distinction and exploitation of homophilic and heterophilic edges can universally benefit representation learning, since two types of edges widely exist in diverse real-world graph data. 
In Table~\ref{tab:main_result_hete}, we find that \ourmethod significantly outperforms conventional and contrastive UGRL methods. The main reason is that these UGRL methods constantly smooth the representations along heterophilic edges, making the representations indistinguishable. In contrast, \ourmethod detects the heterophilic edges and sharpens the representations along them with a high-pass graph filter, leading to {even} {better} performance than semi-supervised GNNs for heterophilic graphs. {Notably, equipping contrastive UGRL methods with heterophily-aware encoders (e.g., GRACE-FA) only yields minor performance gain, indicating that adapting URGL methods to heterophilic graphs needs crafted designs rather than simply modifying the encoder.} \looseness-2

\begin{figure}[!t]
 \centering
 \vspace{-0.2cm}
 \subfigure[Random (Cora)]{
   \includegraphics[width=0.21\textwidth]{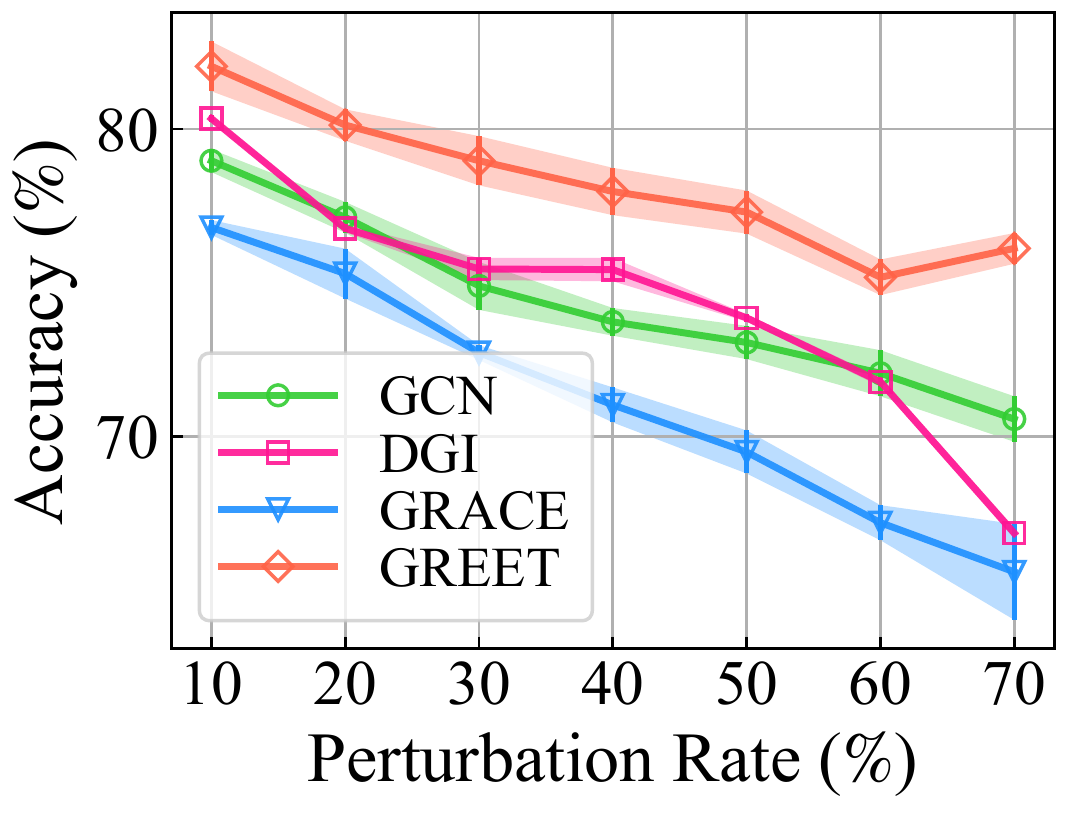}
   \label{subfig:random_cora}
 } 
 \subfigure[Random (Citeseer)]{
   \includegraphics[width=0.21\textwidth]{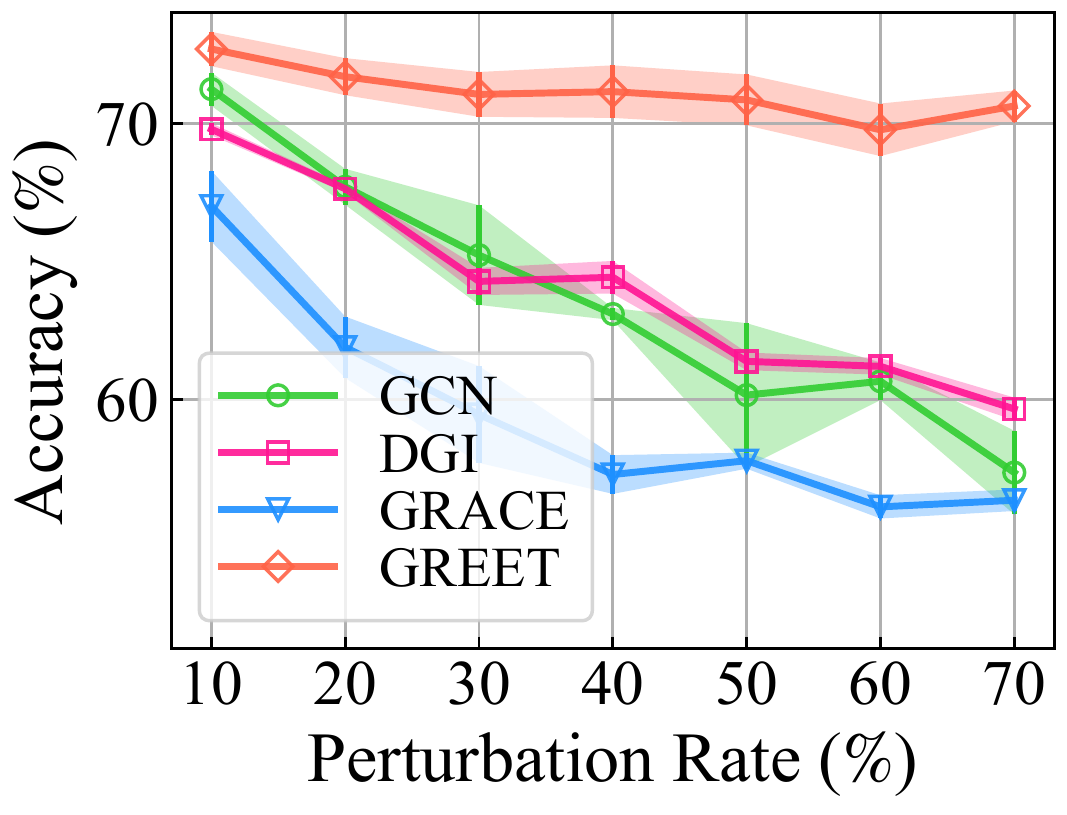}
 }
  \\ 
  \vspace{-0.1cm}
 \subfigure[Metattack (Cora)]{
   \includegraphics[width=0.21\textwidth]{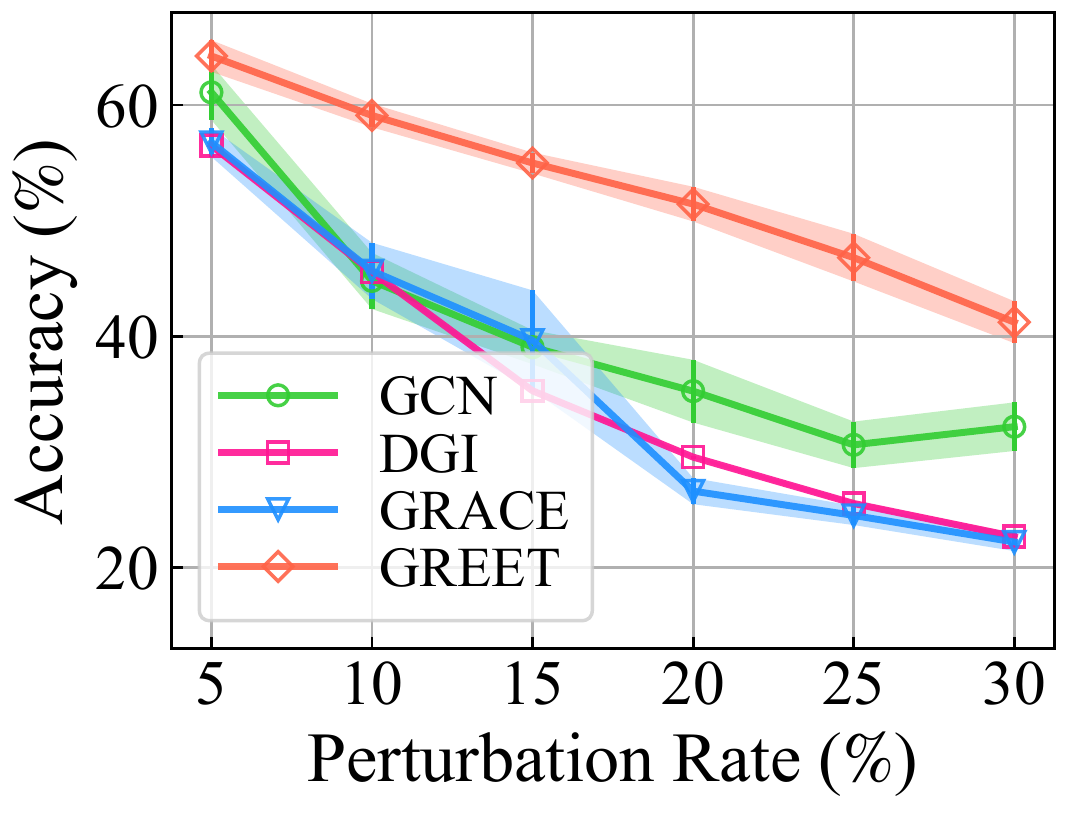}
 }
 \subfigure[Metattack (Citeseer)]{
   \includegraphics[width=0.21\textwidth]{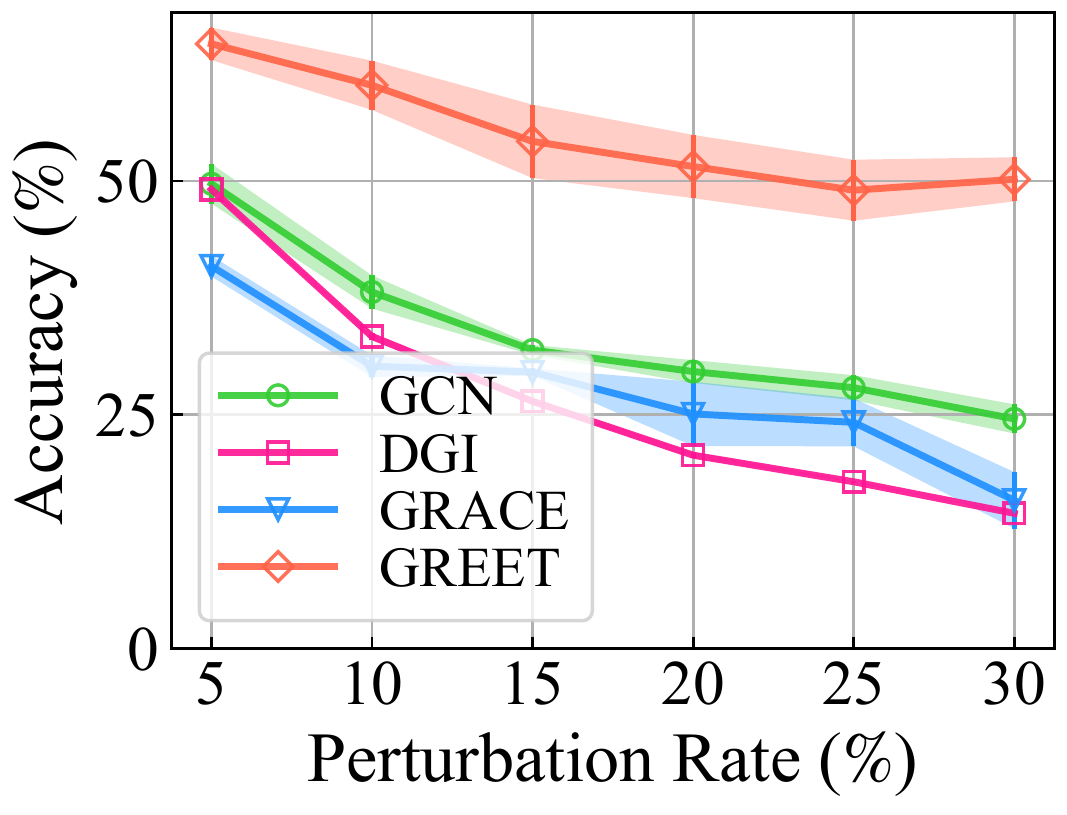}
 }
 \vspace{-0.2cm}
 \caption{Results of different models under random attack (a,b) and Metattack (c,d).}
 \vspace{-0.3cm}
 \label{fig:robust}
\end{figure}

\noindent \textbf{Robustness Analysis. } 
In this experiment, we verify the robustness of \ourmethod on graph data under adversarial attack. We perturb the graph structure with two non-targeted adversarial attack methods, i.e., random attack (randomly adding noisy edges to the original topology) and Metattack~\cite{metattck_zugner2018adversarial}, and test the node classification accuracy on the representations learned from perturbed graphs. The node classification results under different perturbation rates are shown in Fig.~\ref{fig:robust}. From the figure, we can witness that \ourmethod consistently outperforms the baselines by a large margin. Moreover, with the increase of perturbation rate, the advantage of our method becomes more significant. The experimental results demonstrate the strong robustness of \ourmethod against adversarial attack on graph strutures.

\begin{table}[t]
\centering
\caption{Performance of \ourmethod and its variants.} 
\label{tab:ablation}
\vspace{-0.2cm}
\resizebox{1\columnwidth}{!}{
\begin{tabular}{l|cccc}
\toprule
Variants & Cornell & Texas & Cora & CiteSeer   \\
\midrule
\ourmethod & ${85.14{\scriptstyle\pm4.87}}$ & ${87.03{\scriptstyle\pm2.36}}$ & $83.81{\scriptstyle\pm0.87}$ & $73.08{\scriptstyle\pm0.84}$  \\
\midrule
{w/o Eg. Dis.}  & $82.16{\scriptstyle\pm6.64}$ & $82.97{\scriptstyle\pm5.80}$ & $82.19{\scriptstyle\pm0.74}$ & $72.18{\scriptstyle\pm0.58}$ \\
{w/o Du. Enc.} & $77.57{\scriptstyle\pm6.05}$ & $80.54{\scriptstyle\pm4.32}$  & $80.81{\scriptstyle\pm0.54}$ & $72.24{\scriptstyle\pm0.55}$\\
\midrule
{GNN Dis.} & $82.97{\scriptstyle\pm6.05}$ & $82.97{\scriptstyle\pm5.68}$  & $83.62{\scriptstyle\pm0.74}$ & $72.58{\scriptstyle\pm0.93}$ \\
{w/o Pivot} & $79.19{\scriptstyle\pm4.99}$ & $82.16{\scriptstyle\pm5.01}$& $82.84{\scriptstyle\pm0.76}$ & $72.54{\scriptstyle\pm0.69}$  \\
{NCE Loss}  & $77.57{\scriptstyle\pm6.17}$ & $80.00{\scriptstyle\pm6.75}$& $82.48{\scriptstyle\pm0.49}$ & $70.00{\scriptstyle\pm1.18}$ \\
\midrule
{Hom. Rep.} & $75.95{\scriptstyle\pm4.43}$ & $67.57{\scriptstyle\pm4.52}$& $83.16{\scriptstyle\pm0.71}$ & $70.60{\scriptstyle\pm2.04}$  \\
{Het. Rep.}  & $82.97{\scriptstyle\pm5.55}$ & $83.78{\scriptstyle\pm5.67}$& $66.77{\scriptstyle\pm3.02}$ & $68.76{\scriptstyle\pm1.44}$ \\
\bottomrule
\end{tabular}}
\vspace{-4mm}
\end{table}


\noindent \textbf{Ablation Study. } 
To examine the contributions of each component and key design in \ourmethod, we conduct experiments on several variants of \ourmethod and the results are shown in Table~\ref{tab:ablation}. We first remove the key components, i.e. edge discriminator 
(w/o Eg. Dis.) and dual-channel {encoding} (w/o Du. Enc.), to investigate their effects. The results show that both components are critical to \ourmethod, and the dual-channel {encoding} seems to contribute more. 
Then, we discuss the effects of three key designs (i.e., structural encoding {in Eq.(\ref{eq:edge_discriminator})}, {pivot-anchored ranking loss}, and {robust dual-channel} contrastive loss) by replacing them with alternative designs (i.e., {GNN-based discriminator without structural encodings, ranking loss between homophilic and heterophilic edges without pivot, and InfoNCE~\cite{simclr_chen2020simple} loss}) respectively. 
We can witness that \ourmethod consistently outperform three variants, which demonstrates the superiority of the our designs. Finally, we investigate the quality of the representations generated from homophily view (Hom. Rep.) and heterophilic view (Het. Rep.). As expected, the homophilic representations are more effective on homophilic graphs (i.e., Cora and CiteSeer), while the heterophilic representations  {favor} heterophilic graphs (i.e., Cornell and Texas). Jointly considering the representations from two views can produce the best results, because both of them contain critical and distinctive information from different perspectives. \looseness-1

\begin{figure}
 \centering
 \vspace{-0.2cm}
 \subfigure[Cora]{
  \includegraphics[height=0.125\textwidth]{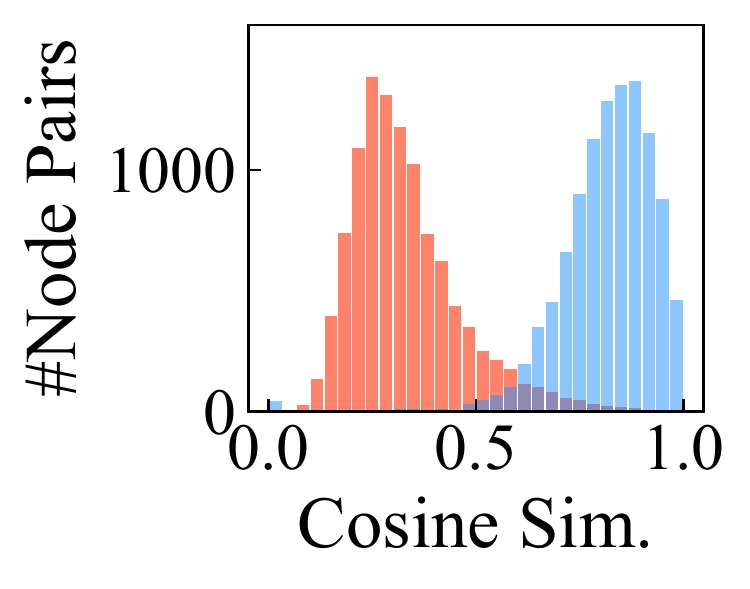}
  \label{subfig:sim_eh_cora}
 } 
 \hspace{-0.33cm}
 \subfigure[Cora-RA]{
  \includegraphics[height=0.125\textwidth]{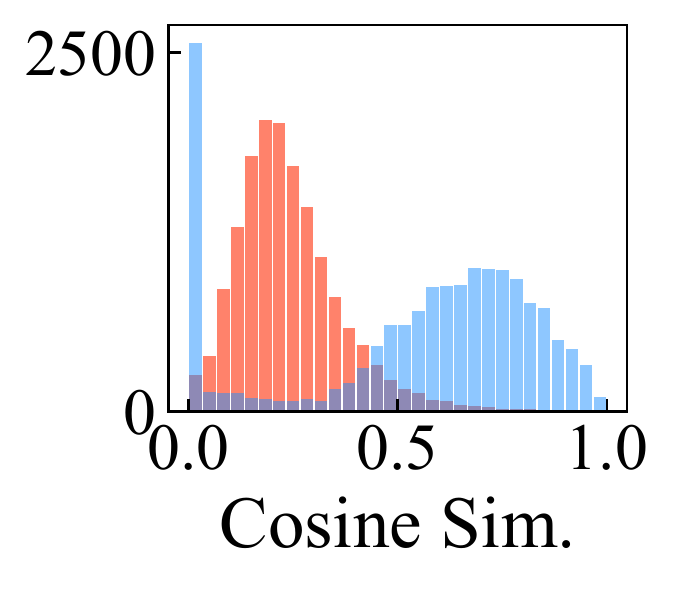}
  \label{subfig:sim_eh_coran}
 }
 \hspace{-0.33cm}
 \subfigure[Texas]{
  \includegraphics[height=0.125\textwidth]{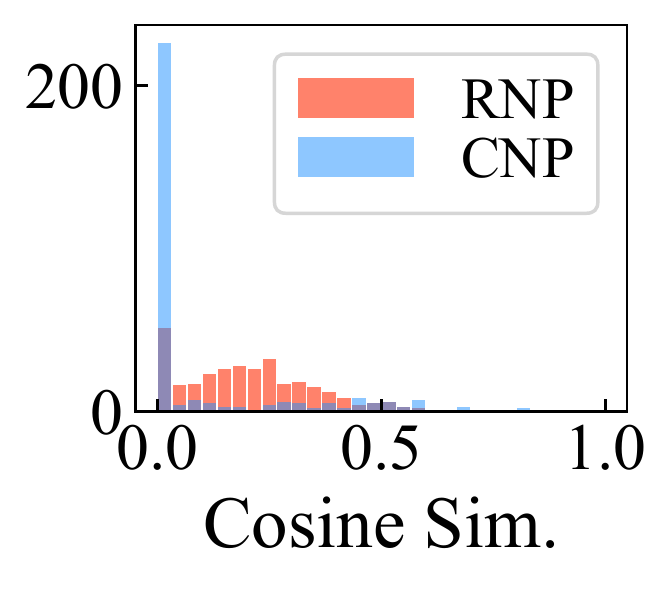}
  \label{subfig:sim_eh_texas}
 }
 \vspace{-0.2cm}
 \caption{The distribution of pair-wise cosine similarity of connected node pairs (CNP) and randomly
sampled node pairs (RNP) w.r.t. the representations learned by \ourmethod on (a) Cora dataset, (b) Cora dataset under random attack with $50\%$ perturbation rate, and (c) Texas dataset.}
 \vspace{-0.4cm}
 \label{fig:sim_exp}
\end{figure}

\noindent \textbf{Visualization. } 
To investigate the property of the representations learned by \ourmethod, we visualize the pair-wise representation similarity of Cora dataset, perturbed Cora dataset, and Texas dataset. 
As shown in Fig.~\ref{subfig:sim_eh_cora}, for most edges in Cora dataset, the representations of end nodes are similar; notably, for a small fraction of edges that are identified to be heterophilic, \ourmethod forces their similarities to be close to $0$. Such observation proves that \ourmethod can separate homophilic and heterophilic edges and generate distinguishable representations accordingly. 
In the perturbed Cora dataset (Fig.~\ref{subfig:sim_eh_coran}), a considerable part of edges are detected to be heterophilic and then have dissimilar end node representations. Thanks to {such a} capability of identifying the noisy edges, \ourmethod shows a strong robustness against structural attack in Fig.~\ref{subfig:random_cora}. 
A similar phenomenon can be found in the heterophilic dataset Texas (Fig.~\ref{subfig:sim_eh_texas}) where the bulk of edges are recognized as heterophilic edges. The separation of two types of edges brings more informative representations, resulting in the superior performance of \ourmethod on heterophilic graphs. \looseness-1

\vspace{-1mm}
\section{Conclusion} \label{sec:conclusion}
In this paper, we propose a novel method named \ourmethod for unsupervised graph representation learning (UGRL). Our core idea is to discriminate and leverage homophilic and heterophilic edges to generate high-quality node representations. 
{We employ an edge discriminator to {distinguish} two types of edges, and construct {a} dual-channel {encoding component} to generate representations according to edge distinction. We carefully design a pivot-anchored ranking loss and robust dual-channel contrastive loss for model {training}, with an alternating {strategy} to train both components in a mutually {boosting} manner. Extensive experiments reveal the effectiveness and robustness of our method.}

\vspace{-1mm}
\section*{Acknowledgements}
The corresponding author is Shirui Pan. 
This work was supported by ARC Future Fellowship (No. FT210100097).

\bibliography{7_reference}

\clearpage

\appendix

\renewcommand\thesubsection{\Roman{section}.\Alph{subsection}}

\section{A. Related Work in Detail} 
In this detailed section on related work, we first introduce two types of conventional GNNs, i.e. spectral-based GNNs and spatial-based GNNs. Then, we review the recent works of heterophily-aware GNNs. After that, we summarize the conventional UGRL methods and contrastive UGRL methods. Finally, we introduce the recent development of multi-view graph representation learning.

\noindent \textbf{Spectral-based GNNs}. 
The spectral-based GNNs perform graph convolution based on graph spectral theory. As a pioneering study of spectral-based GNNs, \cite{estrach2014spectral} first propose the graph convolutional networks based on spectral theory. To further reduce the computational cost, Chebyshev filter~\cite{cheb_defferrard2016convolutional} is applied to spectral-based GNNs for graph convolution. Following these works, GCN~\cite{gcn_kipf2017semi} leverages the first-order approximation of Chebyshev filter as the convolutional operator, which further simplifies the computation. SGC~\cite{sgc_wu2019simplifying} decomposes the graph convolution and feature transformation in spectral-based GNNs, bringing higher running efficiency. More recent works~\cite{hp_dong2021adagnn,hp_ma2021unified} try to improve spectral-based GNNs by introducing adaptive filters or novel convolutional operators. 

\noindent \textbf{Spatial-based GNNs}. 
Following the success of spectral-based GNNs, another type of GNNs, spatial-based GNNs, fastly developed in recent years due to their flexibility. Different from spectral-based GNNs, the spatial-based GNNs perform convolution operation by aggregating local information along edges and then transferring the information (i.e., representations) with learnable neural networks. Following such a message passing paradigm, various spatial-based GNNs with unique designs of aggregation functions are developed. For example, GraphSAGE~\cite{sage_hamilton2017inductive} provides multiple optional aggregation functions for information propagation, including mean pooling, max pooling, and LSTM. GAT~\cite{gat_velivckovic2018graph} introduces self-attention mechanism for aggregation. GIN~\cite{gin_xu2019how} proposes a novel summation aggregation function to increase the expressive power of GNNs. Readers may refer to recent survey~\cite{wu2020comprehensive,zhang2022trustworthy} for a comprehensive review.

\noindent \textbf{Heterophily-aware GNNs}. 
To tackle the challenge of heterophily graph learning, very recently, a family of GNNs focus on learning on graph with heterophilic properties~\cite{zheng2022graph}. For instance, Geom-GCN~\cite{geom_pei2020geom} uses embeddings in a latent space to construct structural neighborhoods for aggregation. FAGCN~\cite{fagcn_bo2021beyond} and GPR-GNN~\cite{gpr_chien2021adaptive} use learnable weights to capture high-frequency patterns. To adapt GCN to graph with low homophily, H2GCN~\cite{h2gcn_zhu2020beyond} incorporates three techniques, i.e., ego- and neighbor- embedding separation, higher-order neighborhoods, and combination of intermediate. UGCN~\cite{ugcn_jin2021universal} extends neighbor set by adding similar but disjunct nodes, such as 2-hop neighboring nodes. DMP~\cite{dmp_yang2021diverse} considers multiple message passing channels to capture different graph signals. WRGNN~\cite{wrgnn2021} considers proximity and structural information to perform multi-relational message aggregation. However, all of these methods require labels to provide supervision signals for representation learning, while how to learn embeddings {for} heterophilic graphs in {the} unsupervised setting still remains {an} open {problem}.

\noindent \textbf{Conventional UGRL}. 
Unsupervised graph representation learning (UGRL, a.k.a. graph embedding) is an essential task in graph machine learning. A line of conventional graph embedding methods uses random walk to construct the proximal neighbors, and then targets to maximize the agreement among the proximal nodes. For instance, Deepwalk~\cite{perozzi2014deepwalk} first builds fix-length random walks and uses a Skip-gram model to learn node embeddings from the walks. 
LINE~\cite{line_tang2015line} considers both first-order and second-order proximity for graph embedding. 
Node2vec~\cite{node2vec_grover2016node2vec} considers breadth-first traversal and
depth-first traversal for random walk generation, and also utilize Skip-gram to learn representations. 
These is also a series of works use anto-encoder-like architecture to maximize the similarities between adjacent nodes. For example, GAE and VGAE~\cite{gae_kipf2016variational} use a binary cross-entropy objective that reconstructs the graph structure to train GNN encoders. ARGA~\cite{arga_pan2018adversarially} further introduces the adversarial mechanism to train the graph auto-encoder model. 

\noindent \textbf{Contrastive UGRL}. 
Following the blossom of contrastive learning~\cite{moco_he2020momentum,simclr_chen2020simple}, some recent efforts apply contrast mechanism to learn representations on graphs in an unsupervised manner~\cite{liu2022graph_ssl_survey,zheng2022rethinking,zheng2022unifying}. In general, these methods construct two augmented views~\cite{ding2022data} from original graphs and then employ contrastive learning objectives (e.g., InfoNCE~\cite{simclr_chen2020simple} or JS Divergence~\cite{dgi_velivckovic2019deep}) to maximize the agreement between corresponding representations {obtained} from two {different} views. For instance, GRACE~\cite{grace_zhu2020deep} use edge dropping and feature masking to generate augmented views, and use InfoNCE for agreement maximization. GCA~\cite{gca_zhu2021graph} further introduces adaptive augmentations to GRACE. GCC~\cite{gcc_qiu2020gcc} establishes graph views by subgraph sampling for contrast. BGRL~\cite{bgrl_thakoor2021large} utilizes a bootstrapping-based learning objective for graph contrastive learning. DGI~\cite{dgi_velivckovic2019deep} and MVGRL~\cite{mvgrl_hassani2020contrastive} introduce a local-global contrastive learning by maximizing the agreement between nodes and full graph with JS Divergence objectives. GraphCL~\cite{graphcl_you2020graph} explores the application of contrastive learning on graph-level learning tasks. Besides UGRL tasks, graph contrastive learning has been proved to be effective on various graph-related tasks, including anomaly detection~\cite{liu2021anomaly}, out-of-distribution detection~\cite{liu2022good}, link prediction~\cite{gmi_peng2020graph}, and federated graph learning~\cite{tan2022fedproto,tan2022federated,tan2023federated}.

\noindent \textbf{Multi-view Graph Representation Learning}. 
Different from the aforementioned methods that learn from a single graph, another line of methods termed multi-view graph representation learning aims to learn representations from multiple {available} graph views~\cite{qu2017attention, dmgi_park2020unsupervised, ma2019multi}. They {learn node representations through fusing} the knowledge from multiple views with {various} mechanisms, {including} attention mechanism~\cite{qu2017attention}, co-regularization~\cite{ni2018co,sun2018multi}, mutual information maximization~\cite{dmgi_park2020unsupervised}, and adversarial training~\cite{fu2020view}. {The representation learning for signed networks has also been explored~\cite{sign3_li2020learning, sign4_lee2020asine},} which can be regarded as a special case of multi-view {graphs} where two {edge} views are {respectively} defined by positive and negative {relations}. The main difference between these methods and our proposed method is that, these methods learn representations from graphs {with multiple explicit} views~\cite{sun2018multi}, while our method {handles} a more challenging task, i.e., learning {to discriminate} homophilic and heterophilic edge views from graphs {with a single edge type} automatically.

\begin{algorithm}[t]
        \caption{Algorithm of \ourmethod}
        {\bf Input:} Edge set $\mathcal{E}$, Feature Matrix $\mathbf{X}$.\\
        {\bf Parameters:} Number of outer iterations $T_o$, Number of inner iterations $T_i$, learning rate of representation learning module $\mu_{grl}$, learning rate of edge discriminator $\mu_{disc}$.\\
        {\bf Output:} Node representation matrix $\mathbf{H}$.\\
     \vspace{-0.3cm}
  \begin{algorithmic}[1]
  \STATE Randomly initialize the parameters of edge discriminator, two encoders and two project heads.
  \FOR{node $v_i \in \mathcal{V}$} 
  \STATE Initialize structural encoding $s_i$ via Eq. (1).
  \STATE Initialize the similar node set $\mathcal{N}_i = \mathrm{kNN}(v_i, k)$.
  \ENDFOR
  \FOR{outer iteration $1,2,\cdots,T_o$} 
  \FOR{inner iteration $1,2,\cdots,T_i$}
  \STATE Calculate $\hat{w}$ with edge discriminator via Eq. (2) and (3).
  \STATE Establish $\mathcal{G}^{(hm)}$ and $\mathcal{G}^{(ht)}$ via Eq. (4).
  \STATE Augment $\mathcal{G}^{(hm)}$ and $\mathcal{G}^{(ht)}$ by edge dropping and feature masking.
  \STATE Calculate $\mathbf{H}^{(hm)}$, $\mathbf{H}^{(ht)}$ via Eq. (5) and (6).
  \STATE Calculate $\mathbf{Z}^{(hm)}$, $\mathbf{Z}^{(ht)}$ with projection heads.
  \STATE Calculate $\mathcal{L}_c$ via Eq. (10).
  \STATE Back-propagate $\mathcal{L}_c$ to update encoders and projection heads with learning rate $\mu_{grl}$.
  \ENDFOR
  \STATE Calculate $\hat{w}$ with edge discriminator via Eq. (2) and (3).
  \STATE Establish $\mathcal{G}^{(hm)}$ and $\mathcal{G}^{(ht)}$ via Eq. (4).
  \STATE Calculate $\mathbf{H}^{(hm)}$, $\mathbf{H}^{(ht)}$ via Eq. (5) and (6).
  \STATE Acquire $\mathbf{H}$ via $\mathbf{H} = [\mathbf{H}^{(hm)}| \mathbf{H}^{(ht)}]$.
  \STATE Calculate $\mathcal{L}_r$ via Eq. (7), (8), and (9).
  \STATE Back-propagate $\mathcal{L}_r$ to update edge generator with learning rate $\mu_{disc}$.
  \ENDFOR
  \STATE Calculate $\hat{w}$ with edge discriminator via Eq. (2) and (3).
  \STATE Establish $\mathcal{G}^{(hm)}$ and $\mathcal{G}^{(ht)}$ via Eq. (4).
  \STATE Calculate $\mathbf{H}^{(hm)}$, $\mathbf{H}^{(ht)}$ via Eq. (5) and (6).
  \RETURN $\mathbf{H} = [\mathbf{H}^{(hm)}| \mathbf{H}^{(ht)}]$
  \end{algorithmic}
  \label{alg:algorithm}
\end{algorithm}

\section{B. Algorithm} \label{appendix:algo}

The training algorithm of \ourmethod is summarized in Algorithm~\ref{alg:algorithm}. As we can see, the structural encodings and similar node sets can be pre-computed in the initialization phase. In each iteration, we first train the graph representation learning module (i.e., dual-channel encoders and projection heads) with the robust dual-channel contrastive loss $\mathcal{L}_c$ for $T_i$ steps ($T_i$ is a hyper-parameter), while keeping the parameters of edge discriminator fixed. In this phase, the learned edge distinction is used to generate graph views, and the contrastive loss trains dual-channel encoders by maximizing the agreement between the current two views. After that, we fix the encoders and learn the edge discriminator for one step under the supervision of pivot-anchored ranking loss $\mathcal{L}_r$. In this phase, the edge discriminator learns better edge distinction according to the latest learned representations. In such an alternating training procedure, the discriminator and encoders are mutually enhanced by each other, and finally, \ourmethod learns the optimal node representations. 

\begin{table*}[t!]
\centering
\caption{Statistics of datasets.}
\begin{tabular}{l|ccccccc}
\toprule
{Dataset} & {Nodes} & {Edges}  & {Classes} & {Features} & $\mathcal{H}_{edge}$ & $\mathcal{H}_{node}$ & {{Train/}{Val/Test}} \\ \midrule
Cora       & 2,708    & 10,556    & 7       & 1,433 & 0.810 & 0.825 & 140 / 500 / 1,000 \\ 
CiteSeer   & 3,327    & 9,104     & 6       & 3,703 & 0.736 & 0.717 & 120 / 500 / 1,000 \\ 
PubMed	   & 19,717   & 88,648    & 3       & 500   & 0.802 & 0.792 & 60 / 500 / 1,000  \\ 
Wiki-CS    & 11,701   & 431,206   & 10      & 300   & 0.654 & 0.677 & 1,170 / 1,171 / 9,360 \\ 
Amz. Comp. & 13,752   & 491,722   & 10      & 767   & 0.777 & 0.802 & 1,375 / 1,376 / 11,001 \\ 
Amz. Photo & 7,650    & 238,162   & 8       & 745   & 0.827 & 0.849 & 765 / 765 / 6,120   \\ 
Co. CS     & 18,333   & 163,788   & 15      & 6,805 & 0.808 & 0.832 & 1,833 / 1,834 / 14,666 \\ 
Co. Physics& 34,493   & 495,924   & 5       & 8,415 & 0.931 & 0.915 & 3,449 / 3,450 / 27,594 \\ 
\midrule
Cornell    & 183      & 295       & 5       & 1,703 & 0.298 & 0.386 & 87 / 59 / 37 \\ 
Texas      & 183      & 309       & 5       & 1,703 & 0.061 & 0.097 & 87 / 59 / 37 \\ 
Wisconsin  & 251      & 499       & 5       & 1,703 & 0.170 & 0.150 & 120 / 80 / 51 \\ 
Chameleon  & 2,277    & 36,051    & 5       & 2,325 & 0.234 & 0.247 & 1,092 / 729 / 456 \\ 
Squirrel   & 5,201    & 216,933   & 5       & 2,089 & 0.223 & 0.216 & 2,496 / 1,664 / 1,041 \\ 
Actor      & 7,600    & 29,926    & 5       & 932   & 0.216 & 0.221 & 3,638 / 2,432 / 1,520 \\ 
\bottomrule
\end{tabular}
\label{tab:dataset}
\end{table*}

\section{C. Complexity Analysis} \label{appendix:complexity}

We discuss time complexity of \ourmethod. 
{In pre-processing phase, the structural encodings and kNN neighbors can be pre-computed at once. Using sparse matrix multiplication, the time complexity for structural encoding computation is $\mathcal{O}(nmd_s)$. Considering the graphs are often sparse ($m \ll n^2$) and $d_s$ can set to be a small value ($d_s \leq 20$), the computational cost here is acceptable. The time complexity of kNN is $\mathcal{O}(n^2)$ for full graph computation; however, using the locality-sensitive kNN approximation algorithm~\cite{fatemi2021slaps,liu2022towards}, this computational cost can be reduce to $\mathcal{O}(nb_k)$ through estimating the top-k neighbors among $b_k$ nodes instead of all nodes.}

{In training phase, the time complexity can be discussed for two main components. }
In edge discriminator, the complexities of the first MLP layer and the second MLP layer are $\mathcal{O}(nd_i(d_f+d_s))$ and $\mathcal{O}(md_i)$ respectively, where $d_i$ is the dimension of intermediate embeddings. 
In dual-channel contrastive learning module, the complexities of encoders and projection heads are $\mathcal{O}(nd_fd_r + md_rL)$ and $\mathcal{O}(nd_rd_pL_p)$, where $L_p$ and $d_p$ are the layer number of projection heads and the dimension of projected latent embeddings, respectively. The complexity of pivot-anchored ranking loss is $\mathcal{O}(md_r)$, while the complexity of robust dual-channel contrastive loss is $\mathcal{O}(nbd_p)$, where $b$ is the batch size of contrastive learning.

\section{D. Datasets} \label{appendix:dataset}

\begin{table*}[t!]
\centering
\caption{Details of the hyper-parameters tuned by grid search.}
\begin{tabular}{l|ccccccccccccc}
\toprule
{Dataset} & $T_o$ & $T_i$ & ${\mu}_{disc}$ & $\alpha$ & $L_p$ & $\gamma^{(hm)}$ & $\gamma^{(ht)}$ & $k$ & $b$ & $p_f^{(hm)}$ & $p_f^{(ht)}$ & $p_e^{(hm)}$ & $p_e^{(ht)}$ \\ \midrule
Cora       & 400  & 2 & 1e-3 & 0.5 & 1 & 0.5 & 0.5 & 20 & FN   & 0.8 & 0.1 & 0.8 & 0.8 \\ 
CiteSeer   & 400  & 2 & 1e-3 & 0.1 & 2 & 0.5 & 0.5 & 30 & FN   & 0.1 & 0.1 & 0.8 & 0.1 \\ 
PubMed	   & 800  & 2 & 1e-3 & 0.1 & 2 & 0.5 & 0.5 & 0  & 5000 & 0.1 & 0.5 & 0.5 & 0.1 \\ 
Wiki-CS    & 1500 & 2 & 1e-3 & 0.1 & 1 & 0.5 & 0.5 & 30 & 3000 & 0.1 & 0.1 & 0.5 & 0.5 \\ 
Amz. Comp. & 1500 & 3 & 1e-4 & 0.3 & 1 & 0.1 & 0.5 & 10 & 5000 & 0.1 & 0.1 & 0.5 & 0.1 \\ 
Amz. Photo & 1500 & 3 & 1e-4 & 0.3 & 1 & 0.5 & 0.5 & 30 & 5000 & 0.1 & 0.1 & 0.8 & 0.5 \\ 
Co. CS     & 1500 & 2 & 1e-3 & 0.5 & 1 & 0.1 & 0.5 & 30 & 5000 & 0.1 & 0.5 & 0.5 & 0.1 \\ 
Co. Physics& 800  & 2 & 1e-3 & 0.1 & 1 & 0.5 & 0.5 & 25 & 2000 & 0.1 & 0.5 & 0.5 & 0.1 \\ 
\midrule
Cornell    & 400  & 2 & 1e-3 & 0.3 & 2 & 0.3 & 0.3 & 25 & FN   & 0.1 & 0.5 & 0.5 & 0.1 \\ 
Texas      & 400  & 2 & 1e-3 & 0.5 & 2 & 0.5 & 0.5 & 20 & FN   & 0.5 & 0.1 & 0.1 & 0.1 \\ 
Wisconsin  & 400  & 2 & 1e-3 & 0.5 & 2 & 0.5 & 0.5 & 25 & FN   & 0.1 & 0.1 & 0.1 & 0.5 \\ 
Chameleon  & 600  & 2 & 1e-3 & 0.1 & 1 & 0.5 & 0.5 & 0  & FN   & 0.1 & 0.5 & 0.5 & 0.1 \\ 
Squirrel   & 1000 & 2 & 1e-3 & 0.1 & 2 & 0.1 & 0.3 & 0  & FN   & 0.1 & 0.1 & 0.1 & 0.8 \\ 
Actor      & 1000 & 2 & 1e-3 & 0.1 & 2 & 0.1 & 0.5 & 20 & FN   & 0.1 & 0.5 & 0.5 & 0.8 \\ 
\bottomrule
\end{tabular}
\label{tab:param_tuned}
\end{table*}

We evaluate our models on eight homophilic graph benchmarks (i.e., Cora, CiteSeer, PubMed, Wiki-CS, Amazon Computer, Amazon Photo, CoAuthor CS, and CoAuthor Physic) and six heterophilic graph benchmarks (i.e., Chameleon, Squirrel, Actor, Cornell, Texas, and Wisconsin). For Cora, CiteSeer, and PubMed datasets, we adopt the public splits~\cite{yang2016revisiting,gcn_kipf2017semi}; for the other 5 homophilic datasets, we adopt a 10\%/10\%/80\% training/validation/testing splits following previous works~\cite{gca_zhu2021graph,bgrl_thakoor2021large}; for heterophilic graphs, we adopt the commonly used 48\%/32\%/20\% training/validation/testing splits used in previous works~\cite{geom_pei2020geom,h2gcn_zhu2020beyond}. The statistics of datasets are provided in Table~\ref{tab:dataset}, where we further attach the edge homophily $\mathcal{H}_{edge}$ \cite{h2gcn_zhu2020beyond} and node homophily $\mathcal{H}_{node}$ \cite{geom_pei2020geom} of each dataset. The details are introduced as follows:

\begin{itemize}
    \item \textbf{Cora, CiteSeer} and \textbf{PubMed} \cite{sen2008collective} are three citation network datasets, where nodes indicate a paper and each edges indicate a citation relationship between two papers. The features are the bag-of-word representations of papers, and labels are the research topic of papers. 
    \item \textbf{Wiki-CS} \cite{mernyei2020wiki} is a reference network extracted from Wikipedia, where nodes indicate articles about computer science and edges indicate the hyperlinks between two articles. The features are the average GloVe word embeddings of the context in the corresponding article, and labels are different fields of each article.
    \item \textbf{Amazon Computers} and \textbf{Amazon Photo} \cite{shchur2018pitfalls} are two  co-purchase networks from Amazon. In these networks, each node indicates a good, and each edge indicates that two goods are frequently bought together. The features is the bag-of-word representations of product reviews, while the labels are the category of goods.
    \item \textbf{CoAuthor CS} and \textbf{CoAuthor Physics} \cite{shchur2018pitfalls} are two co-authorship networks extracted from Microsoft Academic Graph in KDD Cup 2016 challenge. In these networks, nodes indicate an author and edges indicate co-authorship relationships. The features are the bag-of-words embeddings of paper keywords, and the labels are the research fields of authors. 
    \item \textbf{Cornell, Texas} and \textbf{Wisconsin} \cite{geom_pei2020geom} are three web page networks from computer science departments of diverse universities, where nodes are web pages and edges are hyperlinks between two web pages. The features are the bag-of-words representations of the corresponding page, and the labels are types of web pages.
    \item \textbf{Chameleon} and \textbf{Squirrel} \cite{geom_pei2020geom} are two Wikipedia networks where nodes denote web pages in Wikipedia and edges denote links between two pages. The features contain the informative nouns in the Wikipedia pages, and labels stand for the average traffic of the web page.
    \item \textbf{Actor} \cite{geom_pei2020geom} is an actor co-occurrence network where nodes are actors and edges indicate two actors have co-occurrence in the same movie. The features indicate the key word in the Wikipedia pages, and labels are the words of corresponding actors. 
\end{itemize}

\section{E. Implementation Details} \label{appendix:implementation}

\paragraph{Hyper-parameters.}
We select some important hyper-parameters through small grid search, and keep the rest insensitive hyper-parameters to be fixed values. Specifically, the searched hyper-parameters for each benchmark dataset are demonstrated in Table~\ref{tab:param_tuned}, and the selection of fixed hyper-parameters are summarized in Table~\ref{tab:param_fixed}. The grid search is carried out on the following search space:

\begin{itemize}
    \item Number of outer training iterations $T_o$: \{400, 600, 800, 1000, 1500\}
    \item Number of inner training iterations $T_i$: \{1, 2, 3\}
    \item Learning rate of edge discriminator $\mu_{disc}$: \{5e-3, 1e-3, 1e-4\}
    \item High-pass filtering strength $\alpha$: \{0.1, 0.3, 0.5, 0.7\}
    \item Number of projection layers $L_p$: \{1,2\}
    \item Margins of pivot-anchored ranking loss $\gamma^{(hm)}$, $\gamma^{(ht)}$: \{0.1, 0.3, 0.5\}
    \item Number of similar node $k$: \{0, 5, 10, 15, 20, 25, 30\}
    \item Batch size of contrastive loss $b$: \{2000, 3000, 5000, full node (FN)\}
    \item Feature masking rates $p_f^{(hm)}$, $p_f^{(ht)}$: \{0.1 ,0.5, 0.8\}
    \item Edge dropping rates $p_e^{(hm)}$, $p_e^{(ht)}$: \{0.1 ,0.5, 0.8\}
\end{itemize}

\begin{table}[t!]
\centering
\caption{Details of the fixed hyper-parameters.}
\begin{tabular}{l|c}
\toprule
Hyper-parameter & Value \\ \midrule
Learning rate of encoding module $\mu_{grl}$ & 1e-3 \\
Weight decay coefficient $\lambda$           & 0  \\
Dimension of structural encoding $d_s$       & 16 \\
Dimension of intermediate embedding $d_i$    & 128 \\
Dimension of representations $d_r$           & 128 \\
Dimension of projected embedding $d_p$       & 128 \\
Layer number of encoders $L$                 & 2 \\
Temperature of Gumbel-Max $\tau_g$           & 1 \\
Temperature of contrastive loss $\tau_c$     & 0.2 \\

\bottomrule
\end{tabular}
\label{tab:param_fixed}
\end{table}

\paragraph{Evaluation details.} 
We follow previous works \cite{grace_zhu2020deep,gca_zhu2021graph} to construct the downstream classifier for evaluation. To be concrete, we first extract the node representations learned by the UGRL model. Then, the representations of training nodes are used to train an L2-regularized logistic regression classifier from scikit-learn. Finally, the representations of testing nodes are used to evaluate the classification accuracy based on the learned classifier. 

\paragraph{Computing infrastructures.} 
We implement the proposed \ourmethod with PyTorch 1.11.0 and DGL 0.8.0. All experiments are conducted on a Linux server with an Intel Xeon E-2288G CPU and two Quadro RTX 6000 GPUs.

\section{F. Parameter Study} \label{appendix:add_exp}

\begin{figure}[t!]
 \centering
 \subfigure[Sensitivity of $\alpha$]{
   \includegraphics[width=0.22\textwidth]{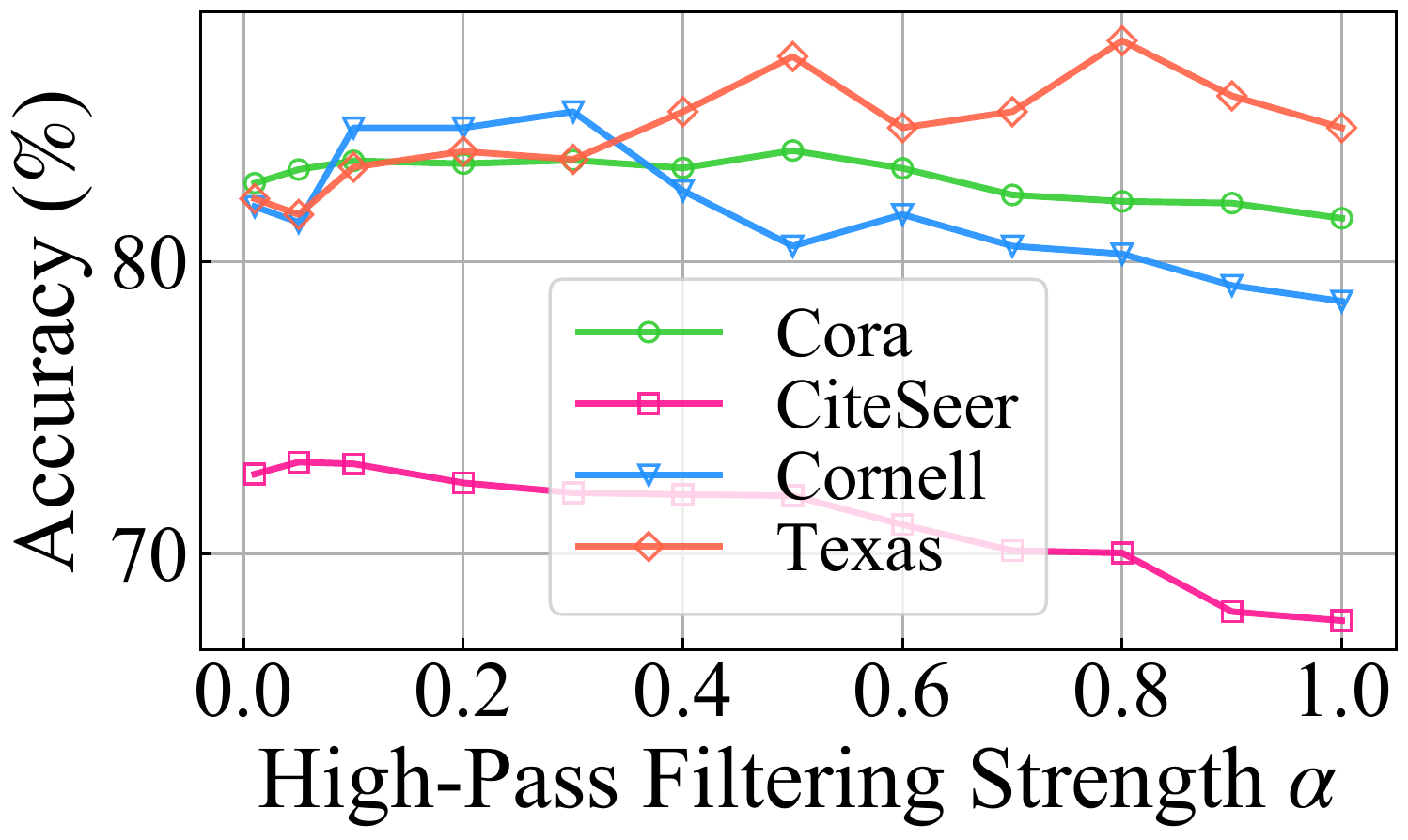}
   \label{subfig:param_alpha}
 } 
 \subfigure[Sensitivity of $k$]{
   \includegraphics[width=0.22\textwidth]{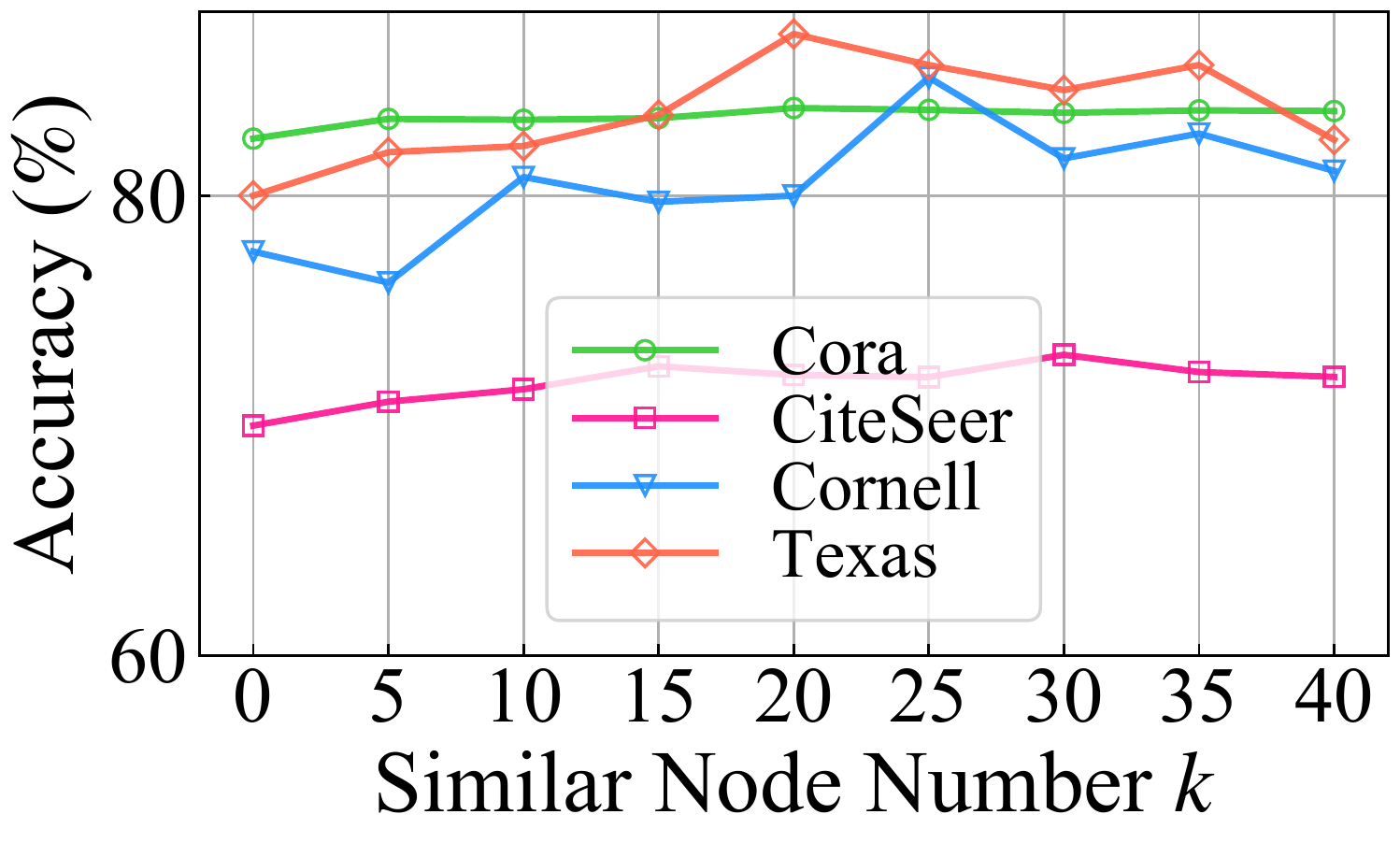}
   \label{subfig:param_k}
 }
 \caption{Parameter sensitivity of $\alpha$ and $k$.}
 \label{fig:param_ak}
\end{figure}

\begin{figure}[t!]
 \centering
 \subfigure[Cora]{
   \includegraphics[width=0.22\textwidth]{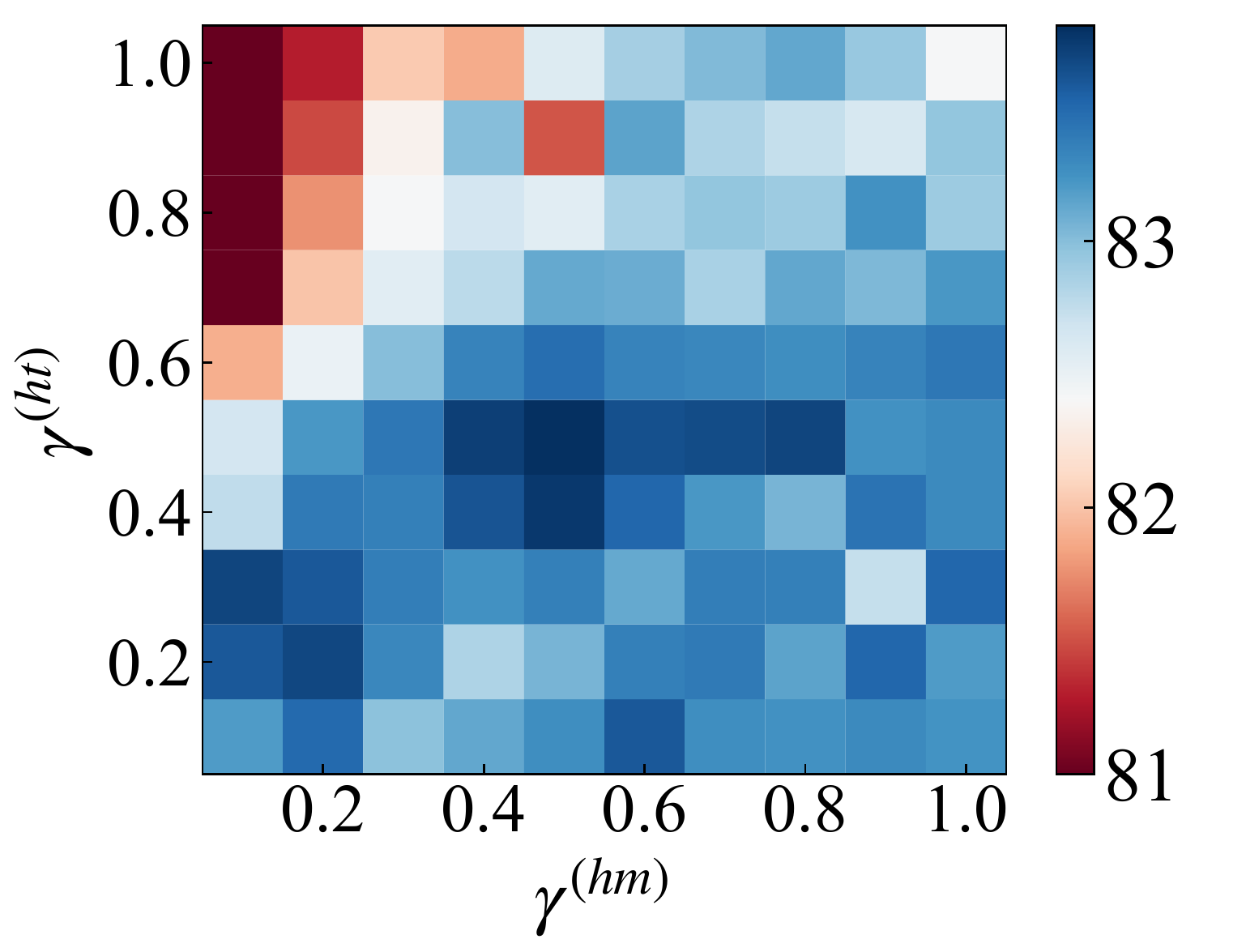}
   \label{subfig:margin_cora}
 } 
 \subfigure[CiteSeer]{
   \includegraphics[width=0.22\textwidth]{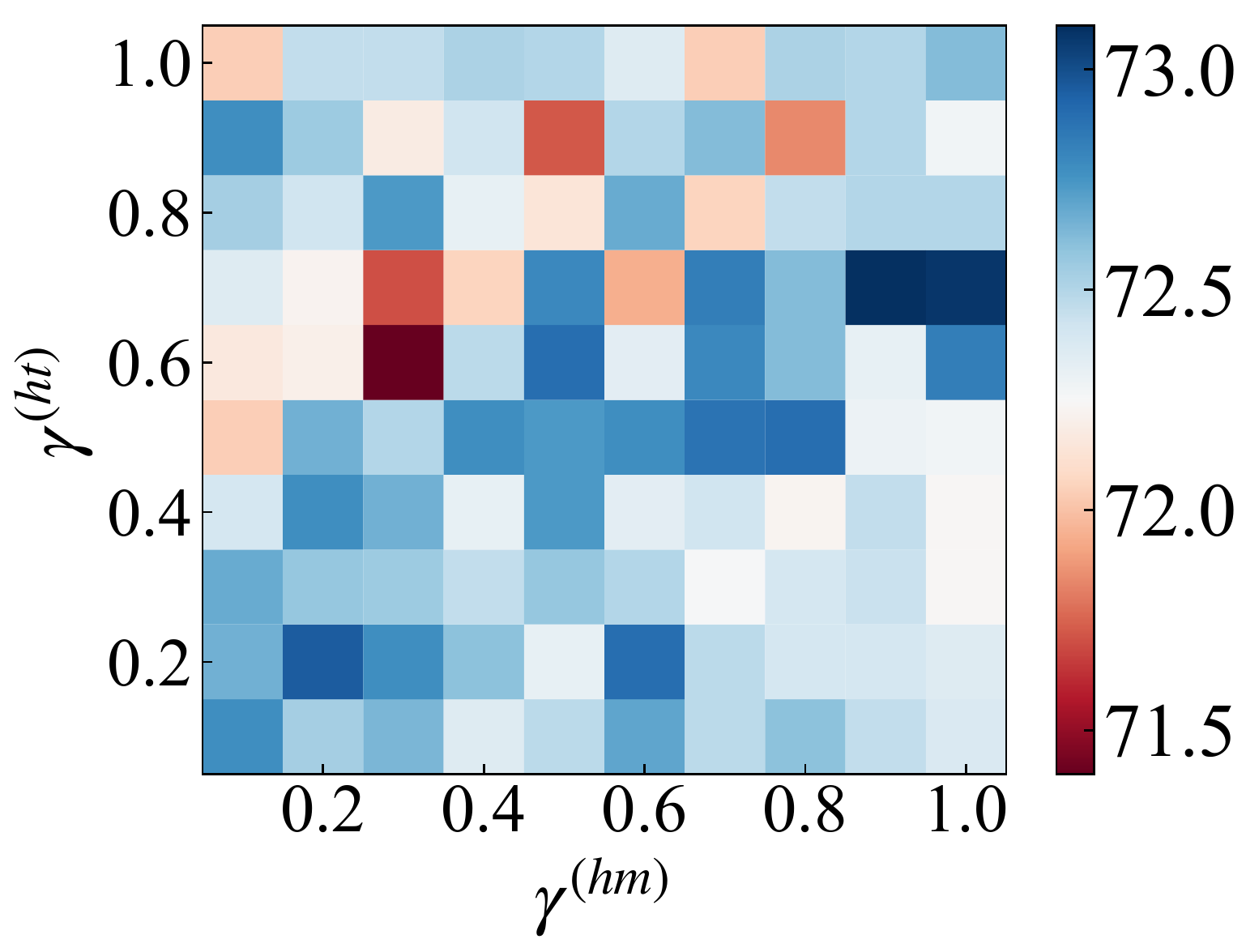}
   \label{subfig:margin_cite}
 } 
  \subfigure[Cornell]{
   \includegraphics[width=0.22\textwidth]{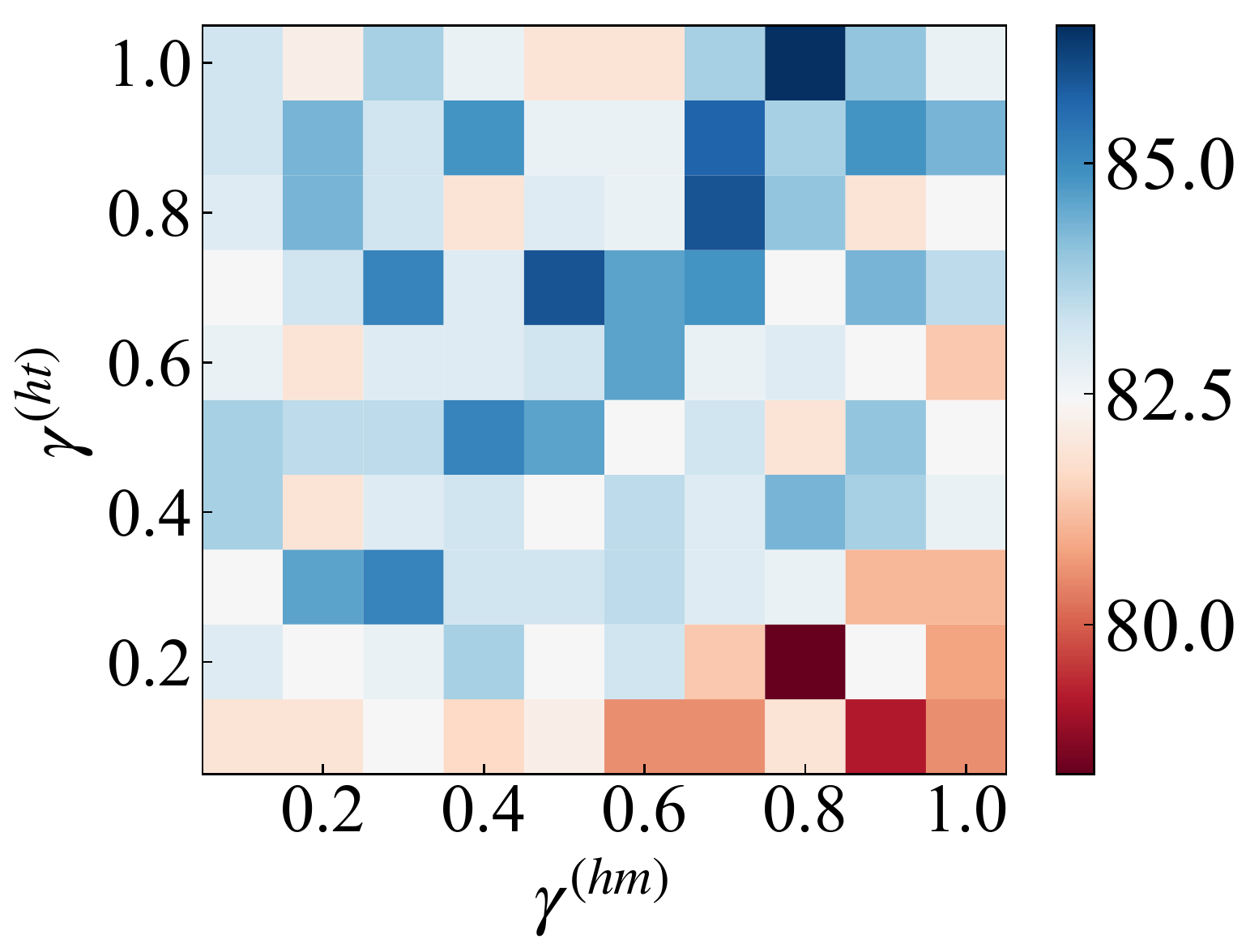}
   \label{subfig:margin_cornell}
 } 
  \subfigure[Texas]{
   \includegraphics[width=0.22\textwidth]{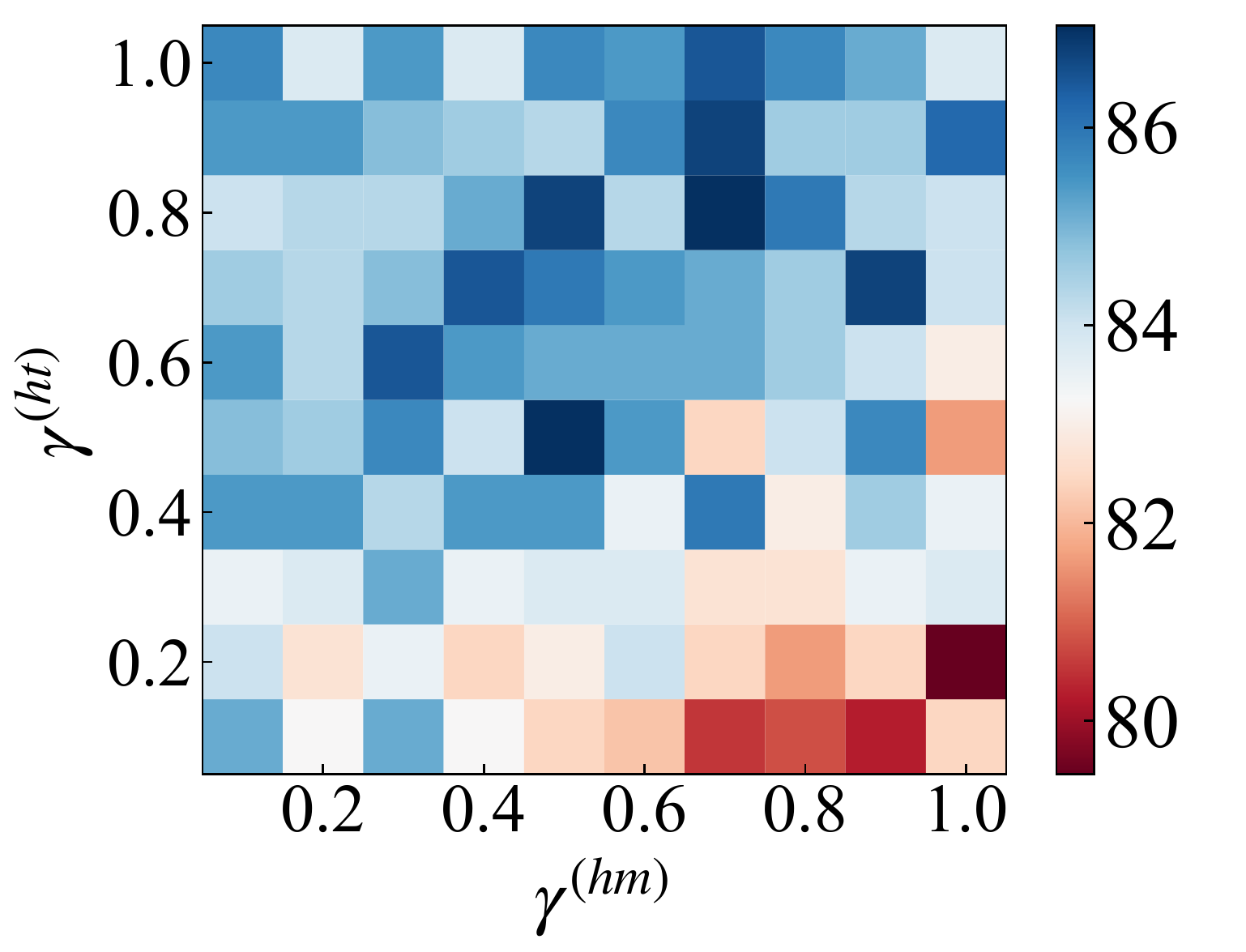}
   \label{subfig:margin_texas}
 } 
 \caption{Parameter sensitivity of $\gamma^{(hm)}$ and $\gamma^{(ht)}$.}
 \vspace{-0.3cm}
 \label{fig:param_margin}
\end{figure}

\begin{table*}[t]
\centering
\caption{Results in terms of classification accuracies (in percent $\pm$ standard deviation) on heterophilic benchmarks. ``*'' indicates that results are borrowed from the original papers. The best and runner-up results are highlighted with \textbf{bold} and \underline{underline}, respectively.} 
\label{tab:main_result_hete_full}
\vspace{-0.1cm}
\resizebox{2\columnwidth}{!}{
\begin{tabular}{p{1.9cm}|p{2.5cm}<{\centering} p{2.5cm}<{\centering} p{2.5cm}<{\centering} p{2.5cm}<{\centering} p{2.5cm}<{\centering} p{2.5cm}<{\centering}}
\toprule
Methods & Chameleon & Squirrel & Actor & Cornell & Texas & Wisconsin \\
\midrule
GCN      & $59.63{\scriptstyle\pm2.32}$ & $36.28{\scriptstyle\pm1.52}$ & $30.83{\scriptstyle\pm0.77}$ & $57.03{\scriptstyle\pm3.30}$ & $60.00{\scriptstyle\pm4.80}$ & $56.47{\scriptstyle\pm6.55}$ \\
GAT      & $56.38{\scriptstyle\pm2.19}$ & $32.09{\scriptstyle\pm3.27}$ & $28.06{\scriptstyle\pm1.48}$ & $59.46{\scriptstyle\pm3.63}$ & $61.62{\scriptstyle\pm3.78}$ & $54.71{\scriptstyle\pm6.87}$ \\
MLP      & $46.91{\scriptstyle\pm2.15}$ & $29.28{\scriptstyle\pm1.33}$ & $35.66{\scriptstyle\pm0.94}$ & $81.08{\scriptstyle\pm7.93}$ & $81.62{\scriptstyle\pm5.51}$ & $84.31{\scriptstyle\pm3.40}$ \\
\midrule
Geom-GCN* & ${60.90}$ & ${38.14}$ & ${31.63}$ & ${60.81}$ & ${67.57}$ & ${64.12}$ \\
H2GCN*    & $59.39{\scriptstyle\pm1.98}$ & $37.90{\scriptstyle\pm2.02}$ & $\underline{35.86{\scriptstyle\pm1.03}}$ & $\underline{82.16{\scriptstyle\pm4.80}}$ & $\underline{84.86{\scriptstyle\pm6.77}}$ & $\mathbf{86.67{\scriptstyle\pm4.69}}$ \\
FAGCN    & $\underline{63.44{\scriptstyle\pm2.05}}$ & $\underline{41.17{\scriptstyle\pm1.94}}$ & $35.74{\scriptstyle\pm0.62}$ & $81.35{\scriptstyle\pm5.05}$ & $84.32{\scriptstyle\pm6.02}$ & $83.33{\scriptstyle\pm2.01}$ \\
GPR-GNN  & $61.58{\scriptstyle\pm2.24}$ & $39.65{\scriptstyle\pm2.81}$ & $35.27{\scriptstyle\pm1.04}$ & $81.89{\scriptstyle\pm5.93}$ & $83.24{\scriptstyle\pm4.95}$ & $84.12{\scriptstyle\pm3.45}$ \\
\midrule
DeepWalk & $47.74{\scriptstyle\pm2.05}$ & $32.93{\scriptstyle\pm1.58}$ & $22.78{\scriptstyle\pm0.64}$ & $39.18{\scriptstyle\pm5.57}$ & $46.49{\scriptstyle\pm6.49}$ & $33.53{\scriptstyle\pm4.92}$ \\
node2vec & $41.93{\scriptstyle\pm3.29}$ & $22.84{\scriptstyle\pm0.72}$ & $28.28{\scriptstyle\pm1.27}$ & $42.94{\scriptstyle\pm7.46}$ & $41.92{\scriptstyle\pm7.76}$ & $37.45{\scriptstyle\pm7.09}$ \\
GAE      & $33.84{\scriptstyle\pm2.77}$ & $28.03{\scriptstyle\pm1.61}$ & $28.03{\scriptstyle\pm1.18}$ & $58.85{\scriptstyle\pm3.21}$ & $58.64{\scriptstyle\pm4.53}$ & $52.55{\scriptstyle\pm3.80}$ \\
VGAE     & $35.22{\scriptstyle\pm2.71}$ & $29.48{\scriptstyle\pm1.48}$ & $26.99{\scriptstyle\pm1.56}$ & $59.19{\scriptstyle\pm4.09}$ & $59.20{\scriptstyle\pm4.26}$ & $56.67{\scriptstyle\pm5.51}$ \\
\midrule
DGI      & $39.95{\scriptstyle\pm1.75}$ & $31.80{\scriptstyle\pm0.77}$ & $29.82{\scriptstyle\pm0.69}$ & $63.35{\scriptstyle\pm4.61}$ & $60.59{\scriptstyle\pm7.56}$ & $55.41{\scriptstyle\pm5.96}$ \\
GMI      & $46.97{\scriptstyle\pm3.43}$ & $30.11{\scriptstyle\pm1.92}$ & $27.82{\scriptstyle\pm0.90}$ & $54.76{\scriptstyle\pm5.06}$ & $50.49{\scriptstyle\pm2.21}$ & $45.98{\scriptstyle\pm2.76}$ \\
MVGRL    & $51.07{\scriptstyle\pm2.68}$ & $35.47{\scriptstyle\pm1.29}$ & $30.02{\scriptstyle\pm0.70}$ & $64.30{\scriptstyle\pm5.43}$ & $62.38{\scriptstyle\pm5.61}$ & $62.37{\scriptstyle\pm4.32}$ \\
GRACE    & $48.05{\scriptstyle\pm1.81}$ & $31.33{\scriptstyle\pm1.22}$ & $29.01{\scriptstyle\pm0.78}$ & $54.86{\scriptstyle\pm6.95}$ & $57.57{\scriptstyle\pm5.68}$ & $50.00{\scriptstyle\pm5.83}$ \\
{GRACE-FA} & $52.68{\scriptstyle\pm2.14}$ & $35.97{\scriptstyle\pm1.20}$ & $32.55{\scriptstyle\pm1.28}$ & $67.57{\scriptstyle\pm4.98}$ & $64.05{\scriptstyle\pm7.46}$ & $63.73{\scriptstyle\pm6.81}$ \\
GCA      & $49.80{\scriptstyle\pm1.81}$ & $35.50{\scriptstyle\pm0.91}$ & $29.65{\scriptstyle\pm1.47}$ & $55.41{\scriptstyle\pm4.56}$ & $59.46{\scriptstyle\pm6.16}$ & $50.78{\scriptstyle\pm4.06}$ \\
BGRL     & $47.46{\scriptstyle\pm2.74}$ & $32.64{\scriptstyle\pm0.78}$ & $29.86{\scriptstyle\pm0.75}$ & $57.30{\scriptstyle\pm5.51}$ & $59.19{\scriptstyle\pm5.85}$ & $52.35{\scriptstyle\pm4.12}$ \\
\midrule 
\ourmethod & $\mathbf{63.64{\scriptstyle\pm1.26}}$ & $\mathbf{42.29{\scriptstyle\pm1.43}}$ & $\mathbf{36.55{\scriptstyle\pm1.01}}$ & $\mathbf{85.14{\scriptstyle\pm4.87}}$ & $\mathbf{87.03{\scriptstyle\pm2.36}}$ & $\underline{84.90{\scriptstyle\pm4.48}}$ \\
\bottomrule
\end{tabular}}
\vspace{-0.2cm}
\end{table*}

We conduct a series of experiments to study the sensitivity of \ourmethod w.r.t. some critical hyper-parameters, including high-Pass filtering strength $\alpha$, contextual node number $k$, and margins of pivot-anchored ranking loss $\gamma^{(hm)}$, $\gamma^{(ht)}$. 

\paragraph{High-Pass Filtering Strength $\alpha$.} In this experiment, we vary $\alpha$ from $0.01$ to $1$ to investigate its effort in \ourmethod. We plot the classification accuracy w.r.t. different selection of $\alpha$ in Fig.~\ref{subfig:param_alpha}. From the figure, we can observe that the best selection of $\alpha$ for different datasets is quiet different. For example, the best selection of $\alpha$ is 0.05 for CiteSeer and $0.8$ for Texas. Besides, a common phenomenon is that the performance consistently decreases when $\alpha$ is too large. A possible reason is that a high-pass filtering with over-large strength would lead to the losing of original graph signals.

\paragraph{Similar Node Number $k$.} We study the sensitivity of \ourmethod w.r.t. the similar node number $k$ by varying $k$ from $0$ to $40$ with an interval of $5$. The experimental results are shown in Fig.~\ref{subfig:param_k}. As we can see in the figure, the performance of \ourmethod drops when $k$ is too large or too small, and the best results on these 4 datasets often occur when $k$ is between $20$ and $30$. We conjecture that an overlarge $k$ would lead to irrelevant contextual nodes, while an extremely small $k$ may result in the lack of supervision signals.

\paragraph{Margins $\gamma^{(hm)}$ and $\gamma^{(ht)}$.} We search the margins for two terms of ranking loss from $0.1$ to $1.0$. Fig. \ref{fig:param_margin} demonstrates the performance under different combinations of $\gamma^{(hm)}$ and $\gamma^{(ht)}$ on four datasets. From the heat maps, we have the following observations. Firstly, \ourmethod is more sensitive to $\gamma^{(ht)}$ rather than $\gamma^{(hm)}$. This observation shows that the constraint on the similarity of heterophilic edges is more important than those of homophilic edges, which demonstrates the significance of detecting heterophilic edges from graph structures. Secondly, for homophilic graph datasets (i.e., Cora and CiteSeer), the performance is better when $\gamma^{(hm)}$ is large and $\gamma^{(ht)}$ is small; for heterophilic graph datasets (i.e., Cornell and Texas), conversely, the performance is better when $\gamma^{(ht)}$ is large and $\gamma^{(hm)}$ is small. A possible reason is that homophilic graphs need tighter constraints for the similarity of homophilic edges since most of the edges are homophilic; similarly, heterophilic graphs require a stricter rule for the similarity of heterophilic edges which account for the majority of edges. Thirdly, when the values of $\gamma^{(hm)}$ and $\gamma^{(ht)}$ are within certain range (i.e., between $0.3$ and $0.7$), \ourmethod consistently performs well on all datasets, which shows that the margins should be set to moderate values. 

\section{G. Detailed Results on Heterophilic Datasets}

The experimental results on 6 heterophilic datasets with standard deviation is demonstrated in Table~\ref{tab:main_result_hete_full}.

\end{document}